\documentclass{article}

\PassOptionsToPackage{numbers, compress}{natbib}

    \usepackage[preprint]{neurips_2023}

\usepackage[hang,flushmargin]{footmisc}
\usepackage{siunitx}
\usepackage{contour}
\usepackage[inline]{enumitem}
\usepackage[utf8]{inputenc} 
\usepackage[T1]{fontenc}    
\usepackage{hyperref}       
\usepackage{url}            
\usepackage{booktabs}       
\usepackage{amsmath,amsfonts,bm}
\usepackage{amssymb}        
\usepackage{nicefrac}       
\usepackage{microtype}      
\usepackage[dvipsnames,table]{xcolor}
\usepackage{graphicx, wrapfig}
\usepackage{epstopdf}
\usepackage[scriptsize,hang]{subfigure}
\usepackage{extarrows}
\usepackage{mathtools,thmtools}
\usepackage{xfrac}
\usepackage{soul}
\usepackage{multirow}
\usepackage{tabularx}
\usepackage{bmpsize}
\usepackage{algorithm}
\usepackage{algorithmic}

\usepackage{times}
\usepackage{epsfig}
\usepackage{array}
\usepackage{collcell}
\usepackage{fp}
\usepackage{xstring}
\usepackage{float}
\usepackage{MnSymbol,bbding,pifont}

\definecolor{color_sn}{HTML}{79a6f6}
\definecolor{color_cheby}{HTML}{a8c0f3}
\definecolor{color_gcn}{HTML}{c8daf4}
\definecolor{color_sgr}{HTML}{e1ebf7}
\definecolor{color_perceptron}{HTML}{ffd6d6}
\definecolor{color_mlp}{HTML}{ffbfbf}
\definecolor{color_filter}{HTML}{fff9db}
\definecolor{color_cnn}{HTML}{fff3ba}
\definecolor{green}{HTML}{1a7a0d}

\definecolor{myred}{HTML}{e53935}
\definecolor{myblue}{HTML}{0277bd}

\definecolor{modelblue}{rgb}{0.00, 0.45, 0.70}

\newcommand{\method}{{\color{modelblue} IGNITE}}
\newcommand{\smethod}{{\color{modelblue} IGNITE }}

\hypersetup{
    colorlinks,
    citecolor=[rgb]{0.9255, 0.0039, 0.5490},
    linkcolor=[rgb]{0, 0, 0},  
    urlcolor=[rgb]{0.00, 0.45, 0.70},  
    }

\usepackage{amsmath,amsfonts,bm}

\def\eqref#1{equation~\ref{#1}}

\def\1{\bm{1}}

\DeclareMathAlphabet{\mathsfit}{\encodingdefault}{\sfdefault}{m}{sl}
\SetMathAlphabet{\mathsfit}{bold}{\encodingdefault}{\sfdefault}{bx}{n}

\graphicspath{{figures/}}

\title{Implicit Geometry and Interaction Embeddings Improve Few-Shot Molecular Property Prediction}

\author{%
  Christopher Fifty$^{\pentagram,1}$, Joseph M. Paggi$^{\pentagram,1}$, Ehsan Amid$^2$, \\
  \textbf{Jure Leskovec}$^1$, \textbf{Ron Dror}$^1$\\
  Equal Contribution$^\pentagram$ \\
  Stanford University$^1$, Google Brain$^2$, \\
  \texttt{fifty@cs.stanford.com} \\
}

\begin{document}

\maketitle

\begin{abstract}

Few-shot learning is a promising approach to molecular property prediction as supervised data is often very limited. However, many important molecular properties depend on complex molecular characteristics --- such as the various 3D geometries a molecule may adopt or the types of chemical interactions it can form --- that are not explicitly encoded in the feature space and must be approximated from low amounts of data. Learning these characteristics can be difficult, especially for few-shot learning algorithms that are designed for fast adaptation to new tasks. 
In this work, we develop molecular embeddings that encode complex molecular characteristics to improve the performance of few-shot molecular property prediction. Our approach leverages large amounts of synthetic data, namely the results of molecular docking calculations, and a multi-task learning paradigm to structure the embedding space.
On multiple molecular property prediction benchmarks, training from the embedding space substantially improves Multi-Task, MAML, and Prototypical Network few-shot learning performance.
Our code is available at \href{https://github.com/cfifty/IGNITE}{https://github.com/cfifty/IGNITE}.

\end{abstract}

\section{Introduction}
In stark contrast to common applications in computer vision and natural language processing, little supervised data is available for many important molecular property prediction tasks~\citep{altae2017low,waring2015analysis}. Determining the biological effects of a molecule---for example, a drug candidate---often involves a lengthy period of cell cultivation followed by a multi-step assay requiring specialized equipment~\citep{lage2018current}. Certain properties, such as toxicity, may even require animal studies to effectively measure~\citep{parasuraman2011toxicological}. The throughput of these assays and studies is limited, and synthesizing molecules to be tested is often expensive~\citep{blakemore2018organic}. As a result, supervised datasets measuring molecular properties often contain on the order of \num{10}s or \num{100}s of data points, and few-shot learning algorithms are necessary to cope with the scarcity of labeled data~\citep{fsmol}.

\begin{figure*}
  \vspace{-0.8cm}
    \centering
    \includegraphics[width=\textwidth]{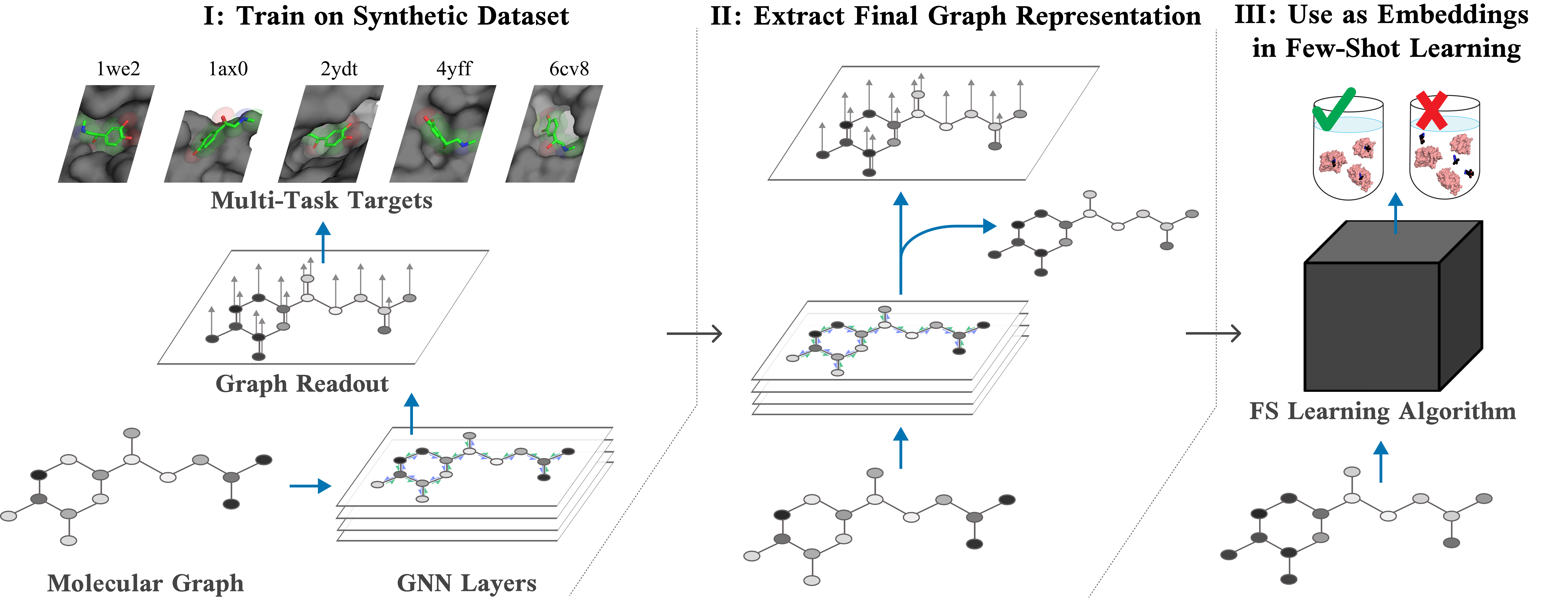}
    \vspace{-0.59cm}
    \caption{\footnotesize{The end-to-end paradigm of developing of \smethod embeddings: (I) train a multi-task GNN to map a small molecule to binding energies (kcal/mol) determined by the interactions between this small molecule and protein targets by physics-based docking. (II) extract the pre-readout graph representation. (III) for a few-shot learning algorithm, replace the default feature space with the learned molecular embeddings. The multi-task targets 1we2, 1ax0, 2ydt, 4yff, and 6cv8, represent Protein Data Bank codes for 5 of the 2,034 targets in our actual dataset.}}
    \label{fig:method}
\vspace{-0.4cm}
\end{figure*}

Few-shot learning algorithms are designed for low-data paradigms, but the effectiveness of these approaches for molecular property prediction, such as the toxicity of a molecule or its binding affinity with a specific protein, is constrained by the inherent complexity of molecular interactions~\citep{chuang2020learning, feinberg2020improvement}. Molecules are specified as a connected graph of atoms, each with its own feature vector, but this feature space does not directly encode complex molecular characteristics, such as the various 3D geometries a molecule may adopt or the types of chemical interactions it can form in each geometry~\citep{yang2019analyzing}. These characteristics are essential to the real-world properties of molecules and deterministic of how molecules interact with other molecules, including proteins, enzymes, biological targets, and anti-targets~\citep{bissantz2010medicinal}.

While molecular representations encoding these attributes may be learned from the feature space, doing so injects an additional layer of complexity into the learning process. As the learning dynamics of few-shot learning algorithms are already quite complex --- and sometimes even unstable~\citep{train_maml} --- training a model to develop both the capacity to rapidly adapt to new tasks as well as to approximate complex molecule-target interactions may be difficult, or even impossible, for existing algorithms. 

In this work, we develop molecular embeddings that encode complex molecular characteristics to improve few-shot molecular property prediction. Our approach begins with curating a large-scale synthetic dataset that measures how \num{1000}s of molecules bind to each one of \num{1000}s of different proteins, therefore implicitly encoding the various 3D geometries a molecule may adopt as well as the chemical interactions it may form within the modeling objective of this dataset. We then train a MPNN~\citep{mpnn} with a multi-task learning scheme---using a single model to predict the interactions among all molecules and targets---to learn a single molecular representation that generalizes to all tasks. Finally, we extract the final-layer learned representation of the multi-task MPNN as a molecular embedding, and use it to initialize the feature space of other few-shot learning algorithms on down-stream molecular property prediction tasks. 
This approach is visualized in \autoref{fig:method}, and because of its ability to capture the implicit characterics of a molecule, we call this method \textbf{I}mplicit \textbf{G}eometry a\textbf{N}d \textbf{I}n\textbf{T}eraction \textbf{E}mbedding (\method). 

In summary, our primary contribution is to develop molecular embeddings that implicitly encode the various 3D geometries a molecule can adopt and the types of chemical interactions it can form, much as word embeddings encode the potential meanings and usages of a word. Our analysis in Section \ref{sec:analysis} shows these characteristics are encoded into the embedding space, so that molecules with similar attributes are close together and molecules with different attributes are far apart. 
Further, on the FS-Mol and MoleculeNet molecular property prediction benchmark suites, with property predictions ranging from toxicity to inhibiting HIV replication, we find that \smethod almost universally improves the performance of Multi-Task, MAML, and Prototypical Network few-shot learning algorithms. This improvement is somewhat remarkable as it is manifest without hyperparameter tuning and by simply modifying the input space upon which the few-shot learning methods operate.

\section{Related Works}
Significant effort has been devoted to addressing challenges associated with training few-shot learning models on small-scale molecular property prediction datasets. We partition these efforts into two collections. The first relates to developing new training objectives for self-supervsed pre-training, while the second focuses on incorporating desirable domain-specific inductive biases into the learning process.

Self-supervised training approaches often aim to learn fundamental representations of small molecules sans labeled data. Many techniques apply a variation of atom, subgraph, or structural motif level masking and reconstruction~\cite{ahmad2022chemberta, pretrain_gnn, mat, rong2020self}, while others begin to incorporate spatial information into molecular graphs to predict the distances among atoms~\cite{pretrain_geom_gnn, jiao2022energy, geomgcl, zaidi2022pre}. Still yet others adopt ideas from contrastive learning: creating augmented views of molecular graphs and then learning graphical representations so that the similarity between different views of the same molecule is maximized while the similarity between views of different molecules is minimized~\cite{hassani2020contrastive, wu2021self, you2020graph, zhu2021graph}. However, \citet{does_gnn_help} finds self-supervised pre-training often conveys a negligible benefit and suggest much of the improvement attributed to self-supervised methods may be derived from extensive hyperparameter tuning of downstream tasks rather than the pre-training strategy itself.

A second perspective focuses on enhancing few-shot learning methodologies with inductive biases desirable for molecular machine learning. \citet{par} propose using a molecule relation graph to improve few-shot learning performance. \citet{mhn} encode context into molecular representations by querying Modern Hopfield Networks to replace a molecule's representation with that of an associated context molecule, and \citet{chen2022meta} augment molecular meta-learning with an adaptive kernel fitting framework. Finally, \citet{fsmol} adapts conventional few-shot learning baselines such as multi-task learning~\cite{mt}, model-agnostic meta-learning~\cite {maml}, and prototypical networks~\cite{proto} to a robust few-shot learning molecular property prediction benchmark. Our findings in Section \ref{sec:experiment} suggest our approach may be complementary to these methods, simply transforming the feature space upon which they operate to an embedding space that is sensitive to complex molecular characteristics such as 3D geometry and chemical interaction.

\section{IGNITE: Implicit Geometry aNd InTeraction Embedding}
While few-shot learning algorithms are designed for low-data paradigms, the effectiveness of these approaches can be limited by the inherent complexity of molecular interactions. We aim to improve their performance by changing the feature space upon which they operate to implicitly encode the various 3D geometries a molecule may adopt as well as the chemical interactions it may form in those geometries. The challenge is to learn an embedding that encodes this information.

\textbf{Synthetic Data Generation.} We may learn such an embedding by pre-training on a sufficiently large amount of data that relies upon 3D molecular geometries and chemical interactions. In principle, this data could be experimental measurements collected in a laboratory that quantify how small molecules bind to proteins. However in practice, experimental data on molecule-protein binding energy is limited, both in quantity and quality. 
Moreover, the measurements in this data come from a wide variety of different assays, from  fluorescence-based functional screens to nuclear magnetic resonance experiments. Comparing the measurements from one assay type with another can be difficult, both for medicinal chemists as well as few-shot learning paradigms.

Instead, we turn to computational ``docking'' methods to generate synthetic data that encodes complex molecular characteristics. In particular, we use Glide, a physics-based docking program~\citep{glide} to curate a large-scale synthetic dataset measuring how \num{1000}s of molecules bind to each of \num{1000}s of different proteins. Using such synthetic data has two key advantages. First, we can generate as much training data as we need. Second, it overcomes the difficulties in comparing experimental measurements across different assay types. 

Given the 3D structure of a target protein and the chemical graph of a small molecule, docking methods estimate the binding energy between the two. Docking matches the process by which real binding energies arise in two key ways. First, docking considers many potential geometries of the protein--small molecule complex, including sampling over the potential internal 3D geometries of the small molecule, and the final prediction is related to the most favorable geometry found. Second, the favorability of each potential geometry is assessed using a physics-inspired scoring function quantifying the complementary of molecular geometries and presence of key chemical interactions. We reason that pre-training on a sufficiently large amount of synthetic data generated by molecular docking would encode molecular attributes deterministic of their interactions into a model's learned representations. To learn these representations, we turn to multi-task learning. 

\textbf{Multi-Task Learning Paradigm.} The synthetic dataset naturally induces a multi-task learning paradigm where each protein represents a separate task. Specifically, for each protein target, we map a small molecule input to its binding energy (kcal/mol) as determined by molecular docking. 
This process is then repeated across the \num{1000}s of protein targets in our dataset to compose a multi-task learning system that uses a single deep learning model to predict how a small molecule will bind to any of the proteins in our dataset. 

With regard to the model itself, we use a 10-layer MPNN~\citep{mpnn} augmented with PNA~\citep{pna} and using a graph readout composed of an element-wise maximum, learned multi-headed weighted sum, and learned multi-headed weighted mean. Each layer in the MPNN uses a Pre-Norm~\citep{xiong2020layer} transformer-like~\citep{attention} residual structure with ReZero weighing~\citep{rezero} using a vector formulation as described in~\citep{deeper}. Moreover, a BOOM-layer~\citep{boom} is used for a second residual connection after message-passing. The post-readout molecular representation is then passed through a shallow MLP that is shared among all tasks before a task-specific linear projection is used to predict the binding energy.

Our optimization protocol uses a linear warmup schedule with 100 steps and Adam~\citep{kingma2014adam} optimizer. The objective function is to minimize the mean squared error between the model's predictions and the free energy measurements. We randomly split our synthetic dataset into 80\% train and 20\% validation, and use early stopping with a window size of 10 epochs on the validation dataset. Training spans 100 epochs, we save the best model checkpoint from early stopping to disk.

\textbf{Usage in Few-Shot Learning.} Many few-shot learning algorithms for molecular property operate on a connected graph of atom-level feature vectors. Atom-level feature vectors often include a one-hot encoding of the element type, atomic charge, mass, valency, etc. 
With \method, we simply change the feature space on which few-shot learning algorithms operate from the default featurization---using a one-hot encoding of element type, atomic charge, etc.---to the pre-readout atom representation learned on the synthetic dataset by  predicting binding energies. 

To summarize, \smethod pre-trains a multi-task model on large-scale synthetic dataset that implicitly encodes complex molecular characteristics---such as the various 3D geometries a molecule may adaopt as well as the chemical interactions it may form---within its modeling objective. It then exports the final MPNN layer's output representation, a connected graph of atom-level learned embeddings, to initialize the feature space of a few-shot learning algorithm. This approach is visualized in \autoref{fig:method}, and our empirical findings in Section~\ref{sec:experiment} suggests it can substantively boost few-shot learning performance.

\section{IGNITE Analysis}
\label{sec:analysis}
While our MPNN model may fit the docking data, it is unclear if relevant biophysical knowledge is actually encoded into the learned \smethod embeddings. Answering this question is the focus of our analysis. 

\subsection{Molecular Space Analysis}
\label{sec:ms_analysis}
Our analysis commences with an exploration of various molecular spaces, and how the molecular space structure implicitly defines a notion of similarity among molecules within that space. 
We present \num{3} ways to measure distance between molecules: docking-based, embedding-based, and fingerprint-based. Later, we use the notion of distance among molecules---distance being a proxy for similarity---to quantify if the molecular characteristics implicitly captured by molecular docking are encoded into the \smethod embedding space. We employ the Kendall Tau measure~\citep{kendall2004course} to objectively quantify this effect.

\textbf{Docking-Based Distance.} Physics-based docking software implicitly encodes a notion of similarity among small molecules. For instance, one may define ``similar'' small molecules as those that manifest similar binding energies across many different protein targets; such molecules typically adopt similar 3D conformations and form similar chemical interactions with the same target. We quantify the implicitly defined similarity between a pair of molecules $m_1$, $m_2$ from docking as the relative difference in docking energies across all targets:
\begin{equation}\label{eqn:1}
    d(m_1, m_2) = \sum\limits_{t \in \text{targets}} \frac{|t(m_1) - t(m_2)|}{\max_{m}\{t(m)\} - \min_{m}\{t(m)\}}\, .
\end{equation}
We compute the relative difference in binding energy as opposed to the absolute difference as different targets have different ranges of binding energies, and we wish to avoid lending more weight to targets with larger ranges than to targets with smaller ranges. 

\textbf{Embedding-Based Distance.} Next we define the distance between molecules in the \smethod embedding space. Our MPNN embedding model uses a combined graph readout composed of an element-wise max, learned weighted sums, and learned weighted means. Applying any of these pooling operations to the per-atom \smethod embeddings will output a component of the vector used to predict the binding energies associated with this target. Accordingly, we choose the simplest operation --- element-wise max pooling --- to motivate the distance between two arbitrary molecules in our embedding space. 

Let $m_1, m_2$ be two molecules parameterized by a per-atom, learned feature matrix and an adjacency matrix in the embedding space: $m_1^{(e)} = (\mathcal{V}^{(e)}_1, \mathcal{E}^{(e)}_1)$. We define $\rho$ to be the max-pooling operation across the embedding dimension for all nodes in the graph. 

For atom embeddings $\mathcal{V}^{(e)} = \begin{bmatrix} v_{1,1} & ... & v_{1,d} \\ ... & & \\ v_{n,1} & ... & v_{n,d}\end{bmatrix} \in \mathbb{R}^{n\times d}$ , 
\begin{equation*}
    \rho(\mathcal{V}^{(e)}) \in \mathbb{R}^{d} ,\,\, \text{where\,\,\,}  [\rho(\mathcal{V}^{(e)})]_j = \max_i v_{ij}\,.
\end{equation*}

We can now define the distance between two molecules in the embedding space as the $l_2$ norm of their element-wise maximum difference:
\begin{equation}\label{eqn:2}
d_{\rho}(m_1, m_2) = \lVert\rho(\mathcal{V}_1) - \rho(\mathcal{V}_2)\rVert_2
\end{equation}

One may also apply \autoref{eqn:2} to the initial feature space used as input to the model:
\begin{equation*}
    m^{(f)}_1 = (\mathcal{V}^{(f)}_1, \mathcal{E}^{(f)}_1)
\end{equation*}
A model's feature space is simply the initial numerical representation of a molecule. Machine learning models often define an atom-level feature space as a numerical vector encoding an atom's element type atomic charge, atomic mass, valency and/or other atomic-level quantities. 

\textbf{Fingerprint-Based Distance.} A third metric of the distance between two molecules is the Tanimoto distance $d_{\tau}$ between their extended-connectivity fingerprints (ECFPs). An ECFP is a  binary string indicating whether or not each of many 2D chemical substructures is present in a given molecule. This Tanimoto distance---which is one of the most commonly used notions of molecular similarity in both academia and industry ~\cite{ecfp}---is defined as the number of chemical sub-structures shared between a pair of molecules divided by the total number of chemical sub-structures present in either molecule. A high value indicates that a pair of molecules share the same components but does not necessarily indicate that those components are connected to one another in the same way or positioned similarly in 3D space.

\textbf{Kendall Tau Measure.} While one may define various metrics to determine the distance among molecules within a molecular space, comparing distances across molecular spaces is more involved. One may abstract the notion of distance between two pairs of molecules to an ordering of all molecules ranked by distance to a single anchor molecule. By ranking the similarity of all molecules to an anchor molecule in one molecular space, we may quantify the extent to which other molecular spaces produce a (dis)similar ordering with respect to this same anchor molecule. 

We employ the Kendall Tau rank distance~\cite{kendall2004course}, also known as the bubble-sort distance, to quantify how similar one ranking of molecular similarity is to the docking-based ordering induced by \autoref{eqn:1}. The Kendall Tau rank distance is given by:
\begin{align*}
K_d(\tau_1, \tau_2) = |(i,j) :& \text{ }i < j, \{[\tau_1(i) \wedge \tau_2(i) > \tau_2(j)] \\
&\vee [\tau_1(i) > \tau_1(j) \wedge \tau_2(i) < \tau_2(j)]\}|
\end{align*}
where $\tau_1(i)$ and $\tau_2(i)$ are the rankings of molecule $i$ given by the rankings $\tau_1$ and $\tau_2$, respectively. Intuitively, Kendall Tau counts the number of swaps made during a bubble sort to convert the ordering in the \smethod embedding space, feature space, or ECFP space to the ordering in the docking space. The smaller the Kendall Tau distance between two molecular spaces with respect to the same anchor molecule, the more ``similar'' they are in the sense of molecular space structure with relation to that anchor molecule. 

\paragraph{Findings.} A naïve Kendall Tau analysis would compare the product space of space of (\num{32547} $\times$ \num{32547}), with each unique molecule in our synthetic dataset serving as an anchor molecule for all other molecules. As this comparison exceeds our computational resources, we randomly sample \num{100} anchor molecules instead to compare the relative orderings induced by various molecule spaces with the ordering induced by the docking-based distance in \autoref{eqn:1}. This analysis approximates how ``similar'' the various molecular spaces are to the docking-based space, the space that directly encodes the various 3D geometries a molecule may adopt as well as the chemical interactions it can form.

Our findings are summarized in \autoref{table:swaps} and indicate ranking molecules by distance in the \smethod embedding space is most similar to the rankings induced by distance in the docking space. Orderings among molecules in ECFP space requires, on average, \num{4.5} million more swaps than orderings in the \smethod embedding space. Similarly, the feature space requires \num{77} million more swaps to reorder its structure to match the docking-space than does the \smethod embedding space. The ``Random'' entry in \autoref{table:swaps} represents a randomly structured molecular space and serves as a reference value for comparison.

\begin{table}[htb!]
    \vspace{-0.2cm}
        \centering
        \small
        \caption{Kendall Tau Analysis. Scores are averaged across 100 randomly selected small molecules. The number of swaps indicates the average number of swaps made by KT to transform an ordering to the docking-based ordering. Normalized Kendall Tau is the number of swaps divided by the number of molecules in an ordering.}
        \vspace{-0.2cm}
        \label{table:swaps}
        \begin{tabular}{l|c|c|}
        \toprule
        Molecular Space & Number of Swaps ({\color{green} \contour{green}{$\downarrow$}}) & Normalized Kendall Tau ({\color{green} \contour{green}{$\downarrow$}}) \\
         \midrule
         Docking & $0$ & $0$ \\
         \smethod & $2.96\times 10^7$ & $0.386$\\
         Feature & $3.73\times 10^7$ & $0.488$\\
         ECFP & $3.28\times 10^7$ & $0.429$\\
         Random & $3.83 \times 10^7$ & $0.50$ \\  
        \bottomrule
        \end{tabular}
    \vspace{-0.6cm}
\end{table}

\subsection{Visualized Example}
\begin{figure}[ht]
  \vspace{-0.2cm}
    \centering
    \includegraphics[width=\textwidth]{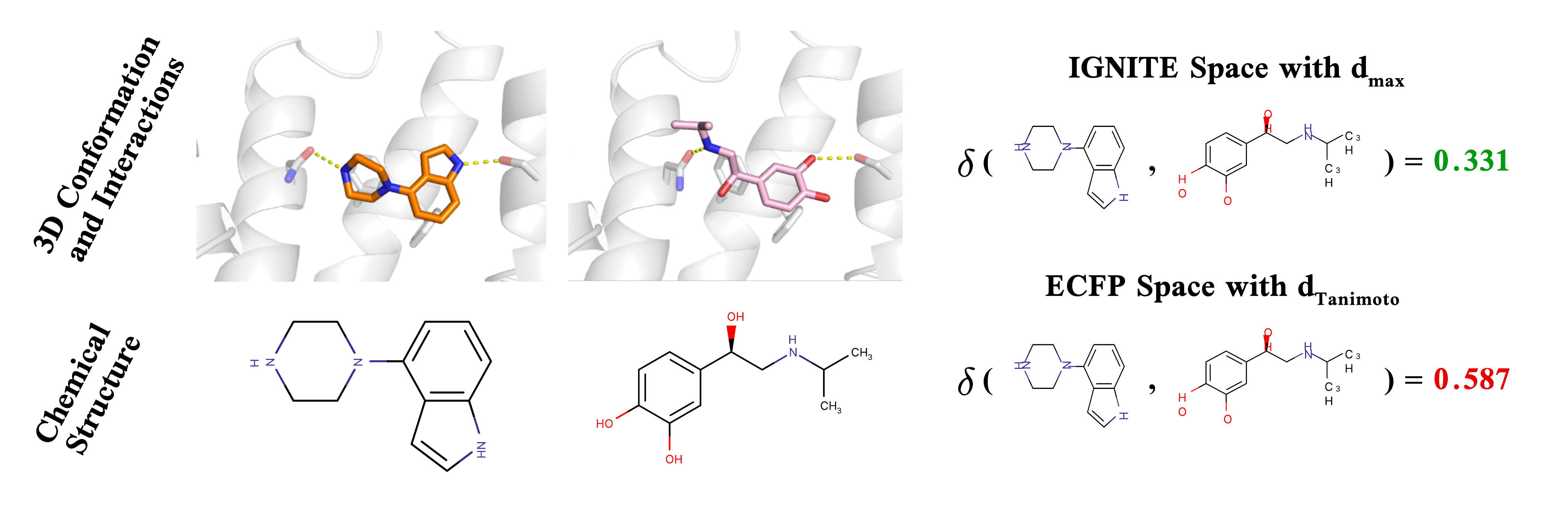}
    \vspace{-0.7cm}
    \caption{\footnotesize{While these two molecules have very different 2D structures, they can adopt similar 3D conformations and form similar interactions with a target protein. Shown are renderings of these molecules in complex with the $\beta_1$-adrenergic receptor, with dashed yellow lines indicating hydrogen bonds. These molecules are relatively close in the embedding space, falling in the 33rd percentile amongst a randomly selected set of molecules, as compared to the 58th percentile using the Tanimoto coefficient in ECFP space.}}
    \label{fig:2}
\vspace{-0.4cm}
\end{figure}

To supplement our quantitative analysis, we offer qualitative analysis of two small molecules that have dissimilar 2D chemical structures, but are known to adopt similar 3D conformations when binding to a biological target.
We select CHEMBL200234 and levisoprenaline as the small molecules and the $\beta_1$-adrenergic receptor (B1AR) as the protein target. While these two molecules are chemically very different, as seen in \autoref{fig:2}, examining the experimentally determined structure of each one bound to B1AR reveals they adopt similar 3D conformations and form similar chemical interactions (e.g., hydrogen bonds) with the target. 

It would be expected that Tanimoto distance in ECFP space determines a low similarity between these two molecules given their disparate chemical structures, but have the complex molecular characteristics shared by these molecules been learned by \method?

To rigorously answer this question, we randomly sample \num{100000} small molecules and order all \num{100002} small molecules (the \num{100000} randomly sampled molecules + CHEMBL200234 and levisoprenaline) with respect to distance from CHEMBL200234 as well as levisoprenaline. We then extract and average the index of levisoprenaline in the ranked list of CHEMBL200234 and the index of CHEMBL200234 in the ranked list of levisoprenaline to determine a combined score. 

Under Tanimoto distance, CHEMBL200234 places levisoprenaline at position \num{50204} while levisoprenaline places CHEMBL200234 at position \num{67119}. In contrast, under the \smethod embeddings distance, CHEMBL200234 places levisoprenaline at position \num{31650} while levisoprenaline places CHEMBL200234 at position \num{34551}. This example---visualized in \autoref{fig:2}---illustrates that molecules even with dissimilar chemical structures may be placed relatively close to one another in the \smethod embedding space. Taken alongside Section~\ref{sec:ms_analysis}, this finding lends support the hypothesis that molecular embeddings learned by pre-training on binding energies implicitly encode complex molecular characteristics.

\section{Experiments and Discussion}
\label{sec:experiment}
Our aim is to determine if \smethod improves few-shot learning performance on molecular property prediction. 
We build our analysis on two conditions reflecting the constraints often imposed on academia and industry where existing models are likely to run efficiently but a capacity --- both human and computational --- to implement new methods and tune them is limited:
\begin{itemize}
    \item \textbf{Model Agnostic}: \smethod should be applicable and effective across a diverse range of architectures and learning algorithms. 
    \item \textbf{Minimal Modification}: replacing the initial feature space with the \smethod embedding space should be effective with minimal modification or hyperparameter tuning. 
\end{itemize}
For our experimental evaluation, we choose FS-Mol~\citep{fsmol}, a few-shot learning benchmark designed to reflect the size and scope of datasets used in drug discovery, as well as \num{3} datasets from the MoleculeNet~\citep{molnet} benchmark suite.

\textbf{FS-Mol Description.} FS-Mol is a few-shot molecular property prediction benchmark that measures how small molecules interact with protein targets. The dataset spans \num{5120} proteins and encompasses \num{233786} different small molecules. On average, each protein has \num{94} small molecule measurements, and the benchmark is formulated to predict if one of these small molecules will inhibit (or not) the protein target. 

The benchmark uses a unique metric, $\Delta$AUPRC, or the relative change in the area under the precision-recall curve (AUPRC) from a random guess baseline. The label distributions of many protein targets in FS-Mol are highly imbalanced; most small molecules will not inhibit a specific protein target. In this setting, measuring the relative change in AUPRC is often more informative than the absolute AUPRC so that targets with balanced labels do not marginalize performance improvements (or deteriorations) of targets with highly imbalanced labels. 

\textbf{FS-Mol Experimental Setup.} \citet{fsmol} presents several few-shot learning baselines for this benchmark. We select three of the best: multi-task pre-training with single-task fine-tuning (MT)~\cite{mt}, model-agnostic meta-learning (MAML)~\cite{maml}, and prototypical networks (PROTO)~\cite{proto}. Our aim is to determine if their performance can be improved by simply modifying the molecular features on which they operate. 

This selection of baselines fulfills our ``Model Agnostic'' criterion as the learning paradigms for MT, MAML, and PROTO are dissimilar. For instance, MT builds molecular representations amenable to all tasks, MAML learns representations that can be quickly adapted, and PROTO leverages a notion of distance---rather than inner product---in its learned representations to predict the label of a molecule. While each baseline is dissimilar, they all use a linear transformation to convert the input feature space to the model's hidden dimension representation space. We simply replace this first hidden dimension representation space with the \smethod embedding space. This change actually simplifies the few-shot learning algorithm, decreasing computational and runtime complexity, by removing the initial embedding layer. 

Adhering to the ``Minimum Modification'' criterion, we do not perform any additional hyperparameter tuning for few-shot learning approaches using \smethod embeddings. We simply use the hyperparameters found by \citet{fsmol} for each baseline model. We prefix few-shot learning methods with an \method- to indicate this model uses \smethod embeddings.

\textbf{FS-Mol Results.} \autoref{table:fsmol_fewshot} indicates that \smethod embeddings significantly improve MT performance on target inhibition prediction, and performance is most improved at smaller support size splits. This result is likely caused from \smethod encoding complex molecular characteristics relevant for protein targets in the test set, characteristics that MT struggles to incorporate into its learned representations from only \num{16} training examples. While the increase in performance declines as the support size increases, it ticks up at \num{256}. This effect is similarly noted by \citet{fsmol} and is caused by few protein targets in the test set actually containing at least \num{256} examples to evaluate the \num{256}-size support split.

We also analyze if \method---an approach developed with a multi-task learning paradigm---might also improve the performance of other few-shot learning algorithms like Prototypical Networks or MAML. This is a difficult proposition as the inductive biases fundamental to each learning algorithm are highly dissimilar. To our surprise, our findings in \autoref{table:fsmol_fewshot} show consistent improvement of PROTO and MAML across all support splits. It may be possible to achieve even larger performance improvements on PROTO and MAML by learning \smethod embeddings with an approach that mirrors the down-steam few-shot learning algorithm. We leave this exploration to future work. 

\begin{table}[t!]
  \vspace{-0.6cm}
  \centering
  \small
  \caption{FS-Mol classification performance across the 157 targets in the FS-Mol test split. We report mean $\Delta$AUPRC and standard error from 10 training runs with different random seeds. Performance of MT, and MAML cited from \cite{fsmol}. PROTO is re-evaluated to use only graph-level features.}
  \vspace{-0.2cm}
  \label{table:fsmol_fewshot}
  \resizebox{\linewidth}{!}{%
  \begin{tabular}{llllll}
    \toprule
      \multirow{2}{*}{Method} & \multicolumn{5}{c}{Support Size}\\ 
      \cmidrule(lr){2-6}
      & \multicolumn{1}{c}{16} & \multicolumn{1}{c}{32} & \multicolumn{1}{c}{64} & \multicolumn{1}{c}{128} & \multicolumn{1}{c}{256}\\
      MT & $0.112 \pm 0.005$ & $0.144 \pm 0.006$ &	$0.177 \pm 0.008$ &	$0.223 \pm 0.008$ & $0.237 \pm 0.018$ \\ 
      \method-MT & $0.157 \pm 0.007$ &	$0.191 \pm 0.008$ &	$0.221 \pm 0.008$ &	$0.259 \pm 0.009$ &	$0.281 \pm 0.020$ \\ 
      \% change & \multicolumn{1}{c}{{\color{green} $\mathbf{\Delta40.2\%}$}} & \multicolumn{1}{c}{{\color{green} $\mathbf{\Delta32.6\%}$}} & \multicolumn{1}{c}{{\color{green} $\mathbf{\Delta24.9\%}$}} & \multicolumn{1}{c}{{\color{green} $\mathbf{\Delta16.1\%}$}} & \multicolumn{1}{c}{{\color{green} $\mathbf{\Delta18.6\%}$}} \\ 
      \midrule 
      MAML & $0.160 \pm 0.008$ & $0.167 \pm 0.008$ & $0.173 \pm 0.008$ & $0.192 \pm 0.009$ & $0.186 \pm 0.019$ \\ 
      \method-MAML & $0.164 \pm 0.008$ & $0.173 \pm 0.008$ & $0.183 \pm 0.008$ & $0.204 \pm 0.008$ & $0.193 \pm 0.018$  \\ 
      \% change & \multicolumn{1}{c}{{\color{green} $\mathbf{\Delta2.5\%}$}} & \multicolumn{1}{c}{{\color{green} $\mathbf{\Delta3.6\%}$}} & \multicolumn{1}{c}{{\color{green} $\mathbf{\Delta5.8\%}$}} & \multicolumn{1}{c}{{\color{green} $\mathbf{\Delta6.3\%}$}} & \multicolumn{1}{c}{{\color{green} $\mathbf{\Delta3.8\%}$}} \\ 
      \midrule 
      PROTO & $0.185 \pm 0.008$ & $0.224 \pm 0.009$ & $0.256 \pm 0.009$ & $0.290 \pm 0.009$ & $0.263 \pm 0.018$ \\ 
      \method-PROTO & $0.195 \pm 0.008$ & $0.232 \pm 0.009$ & $0.263 \pm 0.009$ & $0.297 \pm 0.009$ & $0.283 \pm 0.019$\\ 
      \% change & \multicolumn{1}{c}{{\color{green} $\mathbf{\Delta5.4\%}$}} & \multicolumn{1}{c}{{\color{green} $\mathbf{\Delta3.6\%}$}} & \multicolumn{1}{c}{{\color{green} $\mathbf{\Delta2.7\%}$}} & \multicolumn{1}{c}{{\color{green} $\mathbf{\Delta2.4\%}$}} & \multicolumn{1}{c}{{\color{green} $\mathbf{\Delta7.6\%}$}} \\ 
    \bottomrule
  \end{tabular}
}
\end{table}

\begin{table}[t!]
  \centering
  \small
  \caption{Results on the BACE Classification dataset from the MoleculeNet benchmark suite. We report mean AUPRC as well as standard error averaged across 10 training runs with different random seeds.}
  \vspace{-0.2cm}
  \label{tab:bace}
  \resizebox{\linewidth}{!}{%
  \begin{tabular}{llllll}
    \toprule
      \multirow{2}{*}{Method} & \multicolumn{5}{c}{Support Size}\\ 
      \cmidrule(lr){2-6}
      & \multicolumn{1}{c}{16} & \multicolumn{1}{c}{32} & \multicolumn{1}{c}{64} & \multicolumn{1}{c}{128} & \multicolumn{1}{c}{256}\\
      MT & $0.547 \pm 0.064$ & $0.600 \pm 0.063$ &	$0.652 \pm 0.051$ &	$0.696 \pm 0.037$ & $0.738 \pm 0.028$ \\ 
      \method-MT & $0.574 \pm 0.08$ &	$0.674 \pm 0.039$ &	$0.719 \pm 0.025$ &	$0.739 \pm 0.037$ &	$0.784 \pm 0.014$ \\ 
      \% change & \multicolumn{1}{c}{{\color{green} $\mathbf{\Delta4.9\%}$}} & \multicolumn{1}{c}{{\color{green} $\mathbf{\Delta12.3\%}$}} & \multicolumn{1}{c}{{\color{green} $\mathbf{\Delta10.3\%}$}} & \multicolumn{1}{c}{{\color{green} $\mathbf{\Delta6.2\%}$}} & \multicolumn{1}{c}{{\color{green} $\mathbf{\Delta6.2\%}$}} \\ 
      \midrule 
      MAML & $0.510 \pm 0.026$	& $0.508\pm 0.035$ & $0.548\pm 0.065$ & $0.568 \pm 0.075$ & $0.649 \pm 0.016$ \\ 
      \method-MAML & $0.620\pm 0.027$ & $0.633 \pm 0.031$ & $0.643 \pm 0.043$	& $0.668\pm0.037$ &	$0.685\pm0.041$ \\ 
      \% change & \multicolumn{1}{c}{{\color{green} $\mathbf{\Delta21.6\%}$}} & \multicolumn{1}{c}{{\color{green} $\mathbf{\Delta24.6\%}$}} & \multicolumn{1}{c}{{\color{green} $\mathbf{\Delta17.3\%}$}} & \multicolumn{1}{c}{{\color{green} $\mathbf{\Delta17.6\%}$}} & \multicolumn{1}{c}{{\color{green} $\mathbf{\Delta5.5\%}$}} \\ 
    \bottomrule
    \vspace{-0.6cm}
  \end{tabular}
}
\end{table}

\textbf{MoleculeNet Description \& Setup.} MoleculeNet~\cite{molnet} is a benchmark suite spanning many disparate molecular properties. While each dataset can stand alone, we adapt it to the few-shot learning setting by meta-training on the larger FS-Mol benchmark, sampling a support set from the MoleculeNet dataset, and designating all other examples in this dataset as the query set. As FS-Mol is a binary prediction benchmark, we limit ourselves to binary prediction tasks from the MoleculeNet benchmark and select BACE, HIV, and TOX21 for our experimental analysis. 

BACE measures if a small molecule inhibits (or not) the human $\beta-$secretase-1 enzyme. It contains a single task and measurements for \num{1522} small molecules. Tox21 measures the toxicity of small molecules for several different biological targets such as nuclear receptors and stress pathways. We filter out the targets with high class imbalance, and the resulting dataset contains \num{7831} measures of toxicity across \num{4} targets. HIV measures if a small molecule will inhibit HIV replication. It is composed of a single task and spans \num{41127} small molecules. This dataset is extremely unbalanced, and as both MT and MAML require at least two points from each class to be present---one for training and the other for early stopping when fine-tuning on the support set---we are unable to evaluate HIV at smaller support sizes.

As models are pre-trained on FS-Mol, we use the same fine-tuning hyperparameters and approach as in our earlier evaluation; no additional hyperparameter tuning is performed for \method. However dissimilar to our analysis on FS-Mol, we report AUPRC, or the area under the precision-recall curve, to align our evaluation with the MoleculeNet benchmark~\citep{molnet}. Moreover, due to implementation challenges, we drop our evaluation of PROTO to focus on MT and MAML.

\begin{table}[t!]
  \vspace{-0.8cm}
  \centering
  \small
  \caption{Tox21 classification performance across 4 targets that measure the toxicity of a small molecule with a certain biological pathway or nuclear receptor. We report mean AUPRC as well as standard error averaged across 10 training runs with different random seeds.}
  \vspace{-0.2cm}
  \label{tab:tox21}
  \resizebox{\linewidth}{!}{%
  \begin{tabular}{llllll}
    \toprule
      \multirow{2}{*}{Method} & \multicolumn{5}{c}{Support Size}\\ 
      \cmidrule(lr){2-6}
      & \multicolumn{1}{c}{16} & \multicolumn{1}{c}{32} & \multicolumn{1}{c}{64} & \multicolumn{1}{c}{128} & \multicolumn{1}{c}{256}\\
      MT & $0.120 \pm 0.006$ & $0.140 \pm 0.013$ & $0.155 \pm 0.015$ & $0.166 \pm 0.019$ & $0.198 \pm 0.025$ \\ 
      \method-MT & $0.150 \pm 0.017$ & $0.148 \pm 0.017$ & $0.173 \pm 0.019$ & $0.194 \pm 0.031$ & $0.242 \pm 0.034$ \\ 
      \% change & \multicolumn{1}{c}{{\color{green} $\mathbf{\Delta25.1\%}$}} & \multicolumn{1}{c}{{\color{green} $\mathbf{\Delta5.9\%}$}} & \multicolumn{1}{c}{{\color{green} $\mathbf{\Delta11.7\%}$}} & \multicolumn{1}{c}{{\color{green} $\mathbf{\Delta16.6\%}$}} & \multicolumn{1}{c}{{\color{green} $\mathbf{\Delta22.8\%}$}} \\ 
      \midrule 
      MAML & $0.131 \pm 0.009$ & $0.130 \pm 0.009$ & $0.131 \pm 0.009$ & $0.131 \pm 0.010$ & $0.134 \pm 0.008$ \\ 
      \method-MAML & $0.124 \pm 0.009$ & $0.134 \pm 0.010$ & $0.136 \pm 0.011$ & $0.142 \pm 0.012$ & $0.163 \pm 0.020$ \\ 
      \% change & \multicolumn{1}{c}{{\color{myred} $\mathbf{-\Delta5.2\%}$}} & \multicolumn{1}{c}{{\color{green} $\mathbf{\Delta3.1\%}$}} & \multicolumn{1}{c}{{\color{green} $\mathbf{\Delta3.8\%}$}} & \multicolumn{1}{c}{{\color{green} $\mathbf{\Delta8.2\%}$}} & \multicolumn{1}{c}{{\color{green} $\mathbf{\Delta21.7\%}$}} \\ 
    \bottomrule
    \vspace{-0.6cm}
  \end{tabular}
}
\end{table}

\begin{wraptable}{R}{0.49\textwidth}
  \vspace{-0.6cm}
  \centering
  \small
  \caption{\footnotesize{Results on the HIV dataset from the MoleculeNet benchmark suite. We report mean AUPRC as well as standard error averaged across 10 training runs with different random seeds.}}
  \label{tab:hiv}
  \begin{tabular}{lll}
    \toprule
      \multirow{2}{*}{Method} & \multicolumn{2}{c}{Support Size}\\ 
      \cmidrule(lr){2-3}
      &  \multicolumn{1}{c}{128} & \multicolumn{1}{c}{256}\\
      MT & $0.062\pm0.035$ & $0.078\pm0.033$ \\ 
      \method-MT & $0.066\pm0.029$ & $0.087\pm 0.035$ \\ 
      \% change  & \multicolumn{1}{c}{{\color{green} $\mathbf{\Delta6.5\%}$}} & \multicolumn{1}{c}{{\color{green} $\mathbf{\Delta11.5\%}$}} \\ 
      \midrule
      MAML & $0.053\pm 0.015$ & $0.060\pm 0.023$ \\ 
      \method-MAML  &	$0.070 \pm 0.021$ & $0.068 \pm 0.016$ \\
      \% change  & \multicolumn{1}{c}{{\color{green} $\mathbf{\Delta32.1\%}$}} & \multicolumn{1}{c}{{\color{green} $\mathbf{\Delta13.3\%}$}} \\ 
    \bottomrule
    \vspace{-0.6cm}
  \end{tabular}
\end{wraptable}

\textbf{MoleculeNet Results.} Our results on MoleculeNet datasets reflect our findings on FS-Mol: \smethod appears to almost universally improve few-shot learning performance. Starting with the BACE Classificaiton dataset measuring inhibition of the $\beta$-secretase-1 enzyme, we find augmenting MT and MAML with \smethod embeddings substantively improves performance across all support sizes. Our findings are summarized in \autoref{tab:bace}. However, unlike our findings on FS-Mol, this time MAML benefits the most, improving AUPRC by almost \num{25}\% at \num{32} support size. 

Similar results manifest for the Tox 21 evaluation measuring small molecule toxicities. As shown in \autoref{tab:tox21}, the performance of MT and MAML is improved across all support splits except at the \num{16} support size for MAML. As performance is degraded for the smallest support size, and given the performance improvements at all other support splits, we posit this result may be an outlier caused by an especially detrimental random seed when sampling support-set points for fine-tuning. Smaller support splits are more sensitive to sampling, where the choice of training examples has an outsized influence on model predictions.

The HIV benchmark, measuring if a small molecule inhibits HIV replication, further supports our prior findings that \smethod substantively increases AUPRC for both MT and MAML as shown in \autoref{tab:hiv}. This dataset is especially challenging given its extreme label imbalance. As MT and MAML require at least two points from each class in the support set, we are unable to evaluate this dataset at smaller support splits. However, this setup actually closely mirrors real-life drug discovery where exceedingly few molecules manifest a desired property. It is particularly exciting that \smethod appears to improve the performance of few-shot learning methods on these types of data distributions.

\section{Conclusion}
In this work, we develop molecular embeddings that encode complex molecular characteristics---such as the various 3D geometries a molecule may adopt or the types of chemical interactions it can form---to improve the performance of few-shot learning algorithms on molecular property prediction. 
Our analysis of the \smethod embedding space suggests these characteristics are in fact encoded by the embeddings. Further, our empirical analysis on \num{4} molecular property prediction benchmarks, with properties ranging from the inhibition of protein targets to the toxicity of small molecules, suggests \smethod almost universally improves the performance of \num{3} popular, but very different, few-shot learning algorithms. Moreover, integrating \smethod into few-shot learning algorithms takes minimum modification and does not require additional hyperparameter tuning.

It is our hope that future work will improve upon our approach. Possible avenues for improvement may include leveraging recent advances in generative diffusion models to replace physics-based molecular docking or scaling up to even more synthetic data and expressive architectures.

\newpage

\begin{ack}
We gratefully acknowledge the support of DARPA under Nos. HR00112190039 (TAMI), N660011924033 (MCS); ARO under Nos. W911NF-16-1-0342 (MURI), W911NF-16-1-0171 (DURIP); NSF under Nos. OAC-1835598 (CINES), OAC-1934578 (HDR), CCF-1918940 (Expeditions), NIH under No. 3U54HG010426-04S1 (HuBMAP), Stanford Data Science Initiative, Wu Tsai Neurosciences Institute, Amazon, Docomo, GSK, Hitachi, Intel, JPMorgan Chase, Juniper Networks, KDDI, NEC, and Toshiba.

The content is solely the responsibility of the authors and does not necessarily represent the official views of the funding entities.
\end{ack}

\bibliography{references}
\bibliographystyle{plainnat}

\newpage
{\Large \bf Appendix}
\appendix
\section{GLIDE Protocol}
To produce synthetic data, we used the physics-based docking method GLIDE \cite{glide} (Schrödinger Release 2022-2: Maestro, Schrödinger, LLC, New York, NY, 2022). Running GLIDE can be separated into three steps: preparation of small molecules, preparation of protein structures, and the docking of the prepared small molecules into the prepared protein structures.

To prepare a set of small molecules to be docked, we began with a selection of \num{32547} small molecules from ChEMBL. Importantly, we used ChEMBL simply to provide examples of small molecules; we took the SMILES strings and did not use the annotated activities in any way. We then used the Schrodinger ligprep tool to enumerate relevant tautomeric states (e.g. protonation states) and, if not specified in the SMILES strings, up to 32 stereoisomers.

To prepare a set of protein structures, we began with a set of \num{1601} structures from the PDBBind database. Schrodinger prepwizard was use to add hydrogens, choose tautomeric states, and perform a constrained minimization (heavy atoms within 0.3 Å of starting coordinates) using default parameters \cite{RN10848}. The small molecule annotated in PDBBind was used to center the docking site, but was otherwise not used in any way.

Docking was run using GLIDE in SP mode using default parameters. The docking score for a given small molecule--protein pair was defined as the most favorable score encountered for any tautomeric state of the small molecule, taking into account penalties for choosing an unfavorable tautomeric state. Due to this, it might be that \smethod encodes an implicit representation of a molecule's potential tautomeric states in addition to potential 3D geometries and chemical interactions. Docking calculations were attempted for all small molecule–protein pairs but in some cases docking failed to produce any reasonable poses or the job failed for other reasons so these pairs weren't considered.

\section{Experimental Design}
We offer additional details related to the training of \smethod. While this description can be valuable to some readers, we direct those interested in reproducing our findings or building on our framework to access the code directly. Our code is released in the supplementary material download on OpenReview; however, due to the synethtic data occupying \num{77} GB on disk, we are unable to upload (or link to) the dataset without breaking anonymity. 

\subsection{Additional Training Details}
\begin{wraptable}{R}{0.49\textwidth}
  \centering
  \small
  \caption{\footnotesize{Benchmark comparison.}}
  \label{table:benchmark_comparison}
  \begin{tabular}{lll}
    \toprule
      Dataset & \# tasks & \# compounds\\ 
      \midrule
      FS-Mol Test & $157$ & $27520$ \\ 
      BACE & $1$ & $1522$ \\ 
      Tox21 & $4$ & $7831$ \\ 
      HIV & $1$ & $41127$ \\
    \bottomrule
  \end{tabular}
\end{wraptable}
Our architecture employs a learning rate of 5e-5 for the shared parameters and a learning rate of 1e-4 for the task-specific parameters. We use a linear warm-up scheduler for 100 steps (starting at 0 and ending at the specified learning rate) for both shared and task-specific learning rates. The training process also leverages Adam~\cite{kingma2014adam} with default Pytorch parameters and uses a batch size of 256 molecules. With regards to the task-specific parameters in the \smethod multi-task training paradigm, the output from the graph readout is passed through a shared MLP of hidden dimension \num{512} and then uses a task-specific projection layer of dimension $[512 \times 1]$ for each target. \autoref{table:benchmark_comparison} offers a comparison of the total number of small molecules as well as the number of tasks in each benchmark that we use for evaluation. For our experiments, we used a single \num{80} GB Nvidia A100 GPU, with the model taking approximately \num{10} GB of memory on the GPU. 

\section{Additional FS-Mol Experimental Results}
To offer additional insight into the FS-Mol empirical findings in \autoref{sec:experiment}, we present a per-target breakdown of the first 50 targets in the test set of FS-Mol across support size splits of 16, 32, 64, 128, and 256. \autoref{table:50_mt_fsmol} depicts these results for the Multi-Task Learning baseline, \autoref{table:50_proto_fsmol} depicts these results for the Prototypical Networks baseline, and \autoref{table:50_maml_fsmol} depicts these results from the MAML baseline. Highlighted values indicate the highest result for this target at the given support size.

\begin{table*}[h!]
\centering
\caption{Multi-Task Results measuring $\Delta$AUPRC on the first 50 tasks in the test set of FS-Mol.}
\resizebox{1.0\linewidth}{!}{%
    \begin{tabular}{lllllllllll}
    \toprule
        {TASK-ID} & {16 (\method-MT)} & {32 (\method-MT)} & {64 (\method-MT)} & {128 (\method-MT)} & {256 (\method)} & {16 (MT)} & {32 (MT)} & {64 (MT)} & {128 (MT)} & {256 (MT)} \\
    \midrule
    1006005 & {\color{green} $\mathbf{0.57 \pm 0.05}$} & {\color{green} $\mathbf{0.62 \pm 0.04}$} & {\color{green} $\mathbf{0.63 \pm 0.05}$} & {\color{green} $\mathbf{0.62 \pm 0.05}$} & nan & $0.54 \pm 0.03$ & $0.54 \pm 0.03$ & $0.55 \pm 0.03$ & $0.56 \pm 0.06$ & nan  \\
1066254 & {\color{green} $\mathbf{0.71 \pm 0.07}$} & {\color{green} $\mathbf{0.73 \pm 0.06}$} & {\color{green} $\mathbf{0.81 \pm 0.06}$} & $0.83 \pm 0.10$ & nan & $0.65 \pm 0.05$ & $0.64 \pm 0.09$ & $0.77 \pm 0.07$ & {\color{green} $\mathbf{0.86 \pm 0.11}$} & nan  \\
1119333 & {\color{green} $\mathbf{0.69 \pm 0.08}$} & {\color{green} $\mathbf{0.73 \pm 0.05}$} & {\color{green} $\mathbf{0.78 \pm 0.03}$} & {\color{green} $\mathbf{0.82 \pm 0.03}$} & {\color{green} $\mathbf{0.85 \pm 0.02}$} & $0.69 \pm 0.07$ & $0.72 \pm 0.05$ & $0.76 \pm 0.02$ & $0.79 \pm 0.04$ & $0.84 \pm 0.04$ \\
1243967 & $0.62 \pm 0.08$ & {\color{green} $\mathbf{0.72 \pm 0.09}$} & {\color{green} $\mathbf{0.74 \pm 0.05}$} & {\color{green} $\mathbf{0.80 \pm 0.05}$} & nan & {\color{green} $\mathbf{0.63 \pm 0.07}$} & $0.66 \pm 0.05$ & $0.71 \pm 0.04$ & $0.76 \pm 0.05$ & nan  \\
1243970 & {\color{green} $\mathbf{0.66 \pm 0.06}$} & {\color{green} $\mathbf{0.70 \pm 0.04}$} & {\color{green} $\mathbf{0.75 \pm 0.05}$} & {\color{green} $\mathbf{0.77 \pm 0.04}$} & nan & $0.63 \pm 0.06$ & $0.65 \pm 0.04$ & $0.70 \pm 0.02$ & $0.75 \pm 0.05$ & nan  \\
1613777 & $0.52 \pm 0.03$ & $0.53 \pm 0.04$ & {\color{green} $\mathbf{0.55 \pm 0.02}$} & {\color{green} $\mathbf{0.57 \pm 0.03}$} & {\color{green} $\mathbf{0.59 \pm 0.02}$} & {\color{green} $\mathbf{0.52 \pm 0.04}$} & {\color{green} $\mathbf{0.53 \pm 0.02}$} & $0.54 \pm 0.02$ & $0.57 \pm 0.02$ & $0.59 \pm 0.02$ \\
1613800 & $0.42 \pm 0.02$ & {\color{green} $\mathbf{0.43 \pm 0.03}$} & {\color{green} $\mathbf{0.43 \pm 0.02}$} & $0.45 \pm 0.02$ & $0.47 \pm 0.02$ & {\color{green} $\mathbf{0.42 \pm 0.03}$} & $0.42 \pm 0.01$ & {\color{green} $\mathbf{0.43 \pm 0.03}$} & {\color{green} $\mathbf{0.45 \pm 0.02}$} & {\color{green} $\mathbf{0.47 \pm 0.01}$} \\
1613898 & $0.53 \pm 0.03$ & $0.54 \pm 0.06$ & $0.56 \pm 0.04$ & $0.60 \pm 0.07$ & nan & {\color{green} $\mathbf{0.55 \pm 0.04}$} & {\color{green} $\mathbf{0.54 \pm 0.06}$} & {\color{green} $\mathbf{0.58 \pm 0.04}$} & {\color{green} $\mathbf{0.61 \pm 0.08}$} & nan  \\
1613907 & {\color{green} $\mathbf{0.62 \pm 0.06}$} & {\color{green} $\mathbf{0.62 \pm 0.08}$} & $0.63 \pm 0.08$ & $0.66 \pm 0.14$ & nan & $0.60 \pm 0.05$ & {\color{green} $\mathbf{0.62 \pm 0.07}$} & {\color{green} $\mathbf{0.67 \pm 0.06}$} & {\color{green} $\mathbf{0.70 \pm 0.15}$} & nan  \\
1613926 & {\color{green} $\mathbf{0.67 \pm 0.07}$} & {\color{green} $\mathbf{0.69 \pm 0.06}$} & {\color{green} $\mathbf{0.74 \pm 0.06}$} & {\color{green} $\mathbf{0.85 \pm 0.14}$} & nan & $0.61 \pm 0.04$ & $0.60 \pm 0.06$ & $0.69 \pm 0.06$ & $0.73 \pm 0.19$ & nan  \\
1613949 & {\color{green} $\mathbf{0.47 \pm 0.08}$} & {\color{green} $\mathbf{0.46 \pm 0.06}$} & {\color{green} $\mathbf{0.54 \pm 0.04}$} & {\color{green} $\mathbf{0.62 \pm 0.12}$} & nan & $0.45 \pm 0.06$ & $0.45 \pm 0.06$ & $0.52 \pm 0.06$ & $0.56 \pm 0.12$ & nan  \\
1614027 & {\color{green} $\mathbf{0.54 \pm 0.03}$} & {\color{green} $\mathbf{0.56 \pm 0.04}$} & {\color{green} $\mathbf{0.60 \pm 0.03}$} & {\color{green} $\mathbf{0.66 \pm 0.02}$} & {\color{green} $\mathbf{0.69 \pm 0.03}$} & $0.53 \pm 0.03$ & {\color{green} $\mathbf{0.56 \pm 0.04}$} & $0.59 \pm 0.03$ & $0.63 \pm 0.02$ & $0.66 \pm 0.02$ \\
1614153 & {\color{green} $\mathbf{0.36 \pm 0.01}$} & {\color{green} $\mathbf{0.38 \pm 0.03}$} & {\color{green} $\mathbf{0.38 \pm 0.03}$} & {\color{green} $\mathbf{0.40 \pm 0.02}$} & {\color{green} $\mathbf{0.40 \pm 0.02}$} & $0.36 \pm 0.02$ & $0.37 \pm 0.03$ & $0.37 \pm 0.02$ & $0.38 \pm 0.03$ & $0.39 \pm 0.01$ \\
1614292 & {\color{green} $\mathbf{0.37 \pm 0.03}$} & {\color{green} $\mathbf{0.37 \pm 0.02}$} & {\color{green} $\mathbf{0.38 \pm 0.02}$} & {\color{green} $\mathbf{0.38 \pm 0.02}$} & {\color{green} $\mathbf{0.39 \pm 0.01}$} & $0.36 \pm 0.02$ & $0.37 \pm 0.01$ & $0.38 \pm 0.02$ & $0.38 \pm 0.01$ & $0.38 \pm 0.02$ \\
1614423 & {\color{green} $\mathbf{0.66 \pm 0.12}$} & {\color{green} $\mathbf{0.68 \pm 0.10}$} & {\color{green} $\mathbf{0.72 \pm 0.05}$} & {\color{green} $\mathbf{0.77 \pm 0.04}$} & {\color{green} $\mathbf{0.82 \pm 0.02}$} & $0.52 \pm 0.05$ & $0.54 \pm 0.05$ & $0.59 \pm 0.05$ & $0.67 \pm 0.04$ & $0.74 \pm 0.04$ \\
1614433 & {\color{green} $\mathbf{0.45 \pm 0.04}$} & $0.47 \pm 0.06$ & {\color{green} $\mathbf{0.48 \pm 0.04}$} & {\color{green} $\mathbf{0.50 \pm 0.02}$} & {\color{green} $\mathbf{0.55 \pm 0.04}$} & $0.45 \pm 0.04$ & {\color{green} $\mathbf{0.47 \pm 0.05}$} & $0.47 \pm 0.04$ & $0.49 \pm 0.03$ & $0.53 \pm 0.04$ \\
1614466 & {\color{green} $\mathbf{0.47 \pm 0.04}$} & {\color{green} $\mathbf{0.48 \pm 0.05}$} & {\color{green} $\mathbf{0.49 \pm 0.03}$} & {\color{green} $\mathbf{0.49 \pm 0.05}$} & {\color{green} $\mathbf{0.50 \pm 0.03}$} & $0.46 \pm 0.05$ & $0.47 \pm 0.05$ & $0.47 \pm 0.04$ & $0.48 \pm 0.04$ & $0.48 \pm 0.02$ \\
1614503 & {\color{green} $\mathbf{0.47 \pm 0.08}$} & {\color{green} $\mathbf{0.48 \pm 0.09}$} & {\color{green} $\mathbf{0.52 \pm 0.05}$} & {\color{green} $\mathbf{0.73 \pm 0.21}$} & nan & $0.44 \pm 0.05$ & $0.44 \pm 0.06$ & $0.48 \pm 0.08$ & $0.66 \pm 0.22$ & nan  \\
1614508 & $0.74 \pm 0.10$ & {\color{green} $\mathbf{0.85 \pm 0.02}$} & $0.87 \pm 0.02$ & $0.90 \pm 0.04$ & nan & {\color{green} $\mathbf{0.76 \pm 0.07}$} & {\color{green} $\mathbf{0.85 \pm 0.04}$} & {\color{green} $\mathbf{0.87 \pm 0.03}$} & {\color{green} $\mathbf{0.91 \pm 0.05}$} & nan  \\
1614522 & {\color{green} $\mathbf{0.58 \pm 0.04}$} & {\color{green} $\mathbf{0.61 \pm 0.04}$} & {\color{green} $\mathbf{0.62 \pm 0.03}$} & {\color{green} $\mathbf{0.66 \pm 0.01}$} & {\color{green} $\mathbf{0.66 \pm 0.03}$} & $0.54 \pm 0.04$ & $0.55 \pm 0.02$ & $0.56 \pm 0.02$ & $0.58 \pm 0.03$ & $0.60 \pm 0.02$ \\
1737951 & {\color{green} $\mathbf{0.64 \pm 0.10}$} & {\color{green} $\mathbf{0.67 \pm 0.06}$} & {\color{green} $\mathbf{0.79 \pm 0.05}$} & {\color{green} $\mathbf{0.85 \pm 0.08}$} & nan & $0.55 \pm 0.06$ & $0.60 \pm 0.08$ & $0.63 \pm 0.05$ & $0.72 \pm 0.07$ & nan  \\
1738079 & $0.50 \pm 0.02$ & $0.50 \pm 0.04$ & $0.49 \pm 0.03$ & $0.48 \pm 0.03$ & nan & {\color{green} $\mathbf{0.52 \pm 0.04}$} & {\color{green} $\mathbf{0.52 \pm 0.04}$} & {\color{green} $\mathbf{0.53 \pm 0.03}$} & {\color{green} $\mathbf{0.56 \pm 0.05}$} & nan  \\
1738362 & $0.50 \pm 0.10$ & $0.46 \pm 0.08$ & $0.59 \pm 0.09$ & $0.75 \pm 0.20$ & nan & {\color{green} $\mathbf{0.52 \pm 0.07}$} & {\color{green} $\mathbf{0.56 \pm 0.11}$} & {\color{green} $\mathbf{0.62 \pm 0.05}$} & {\color{green} $\mathbf{0.84 \pm 0.14}$} & nan  \\
1738395 & {\color{green} $\mathbf{0.51 \pm 0.04}$} & {\color{green} $\mathbf{0.49 \pm 0.05}$} & {\color{green} $\mathbf{0.50 \pm 0.04}$} & {\color{green} $\mathbf{0.54 \pm 0.04}$} & nan & $0.49 \pm 0.04$ & $0.48 \pm 0.03$ & $0.50 \pm 0.05$ & $0.51 \pm 0.05$ & nan  \\
1738485 & {\color{green} $\mathbf{0.56 \pm 0.02}$} & $0.55 \pm 0.03$ & {\color{green} $\mathbf{0.58 \pm 0.06}$} & $0.59 \pm 0.04$ & $0.65 \pm 0.06$ & $0.54 \pm 0.04$ & {\color{green} $\mathbf{0.56 \pm 0.03}$} & $0.58 \pm 0.04$ & {\color{green} $\mathbf{0.60 \pm 0.06}$} & {\color{green} $\mathbf{0.68 \pm 0.06}$} \\
1738502 & {\color{green} $\mathbf{0.47 \pm 0.06}$} & {\color{green} $\mathbf{0.50 \pm 0.06}$} & {\color{green} $\mathbf{0.53 \pm 0.05}$} & {\color{green} $\mathbf{0.53 \pm 0.03}$} & {\color{green} $\mathbf{0.58 \pm 0.03}$} & $0.37 \pm 0.03$ & $0.39 \pm 0.02$ & $0.42 \pm 0.03$ & $0.45 \pm 0.02$ & $0.49 \pm 0.04$ \\
1738573 & $0.51 \pm 0.02$ & $0.52 \pm 0.03$ & {\color{green} $\mathbf{0.54 \pm 0.03}$} & {\color{green} $\mathbf{0.55 \pm 0.02}$} & {\color{green} $\mathbf{0.57 \pm 0.01}$} & {\color{green} $\mathbf{0.52 \pm 0.02}$} & {\color{green} $\mathbf{0.53 \pm 0.01}$} & $0.53 \pm 0.01$ & $0.54 \pm 0.01$ & $0.56 \pm 0.02$ \\
1738579 & {\color{green} $\mathbf{0.58 \pm 0.05}$} & {\color{green} $\mathbf{0.59 \pm 0.05}$} & {\color{green} $\mathbf{0.59 \pm 0.04}$} & {\color{green} $\mathbf{0.67 \pm 0.04}$} & nan & $0.52 \pm 0.03$ & $0.54 \pm 0.04$ & $0.56 \pm 0.03$ & $0.62 \pm 0.04$ & nan  \\
1738633 & {\color{green} $\mathbf{0.59 \pm 0.06}$} & {\color{green} $\mathbf{0.68 \pm 0.04}$} & {\color{green} $\mathbf{0.71 \pm 0.06}$} & $0.70 \pm 0.08$ & nan & $0.57 \pm 0.05$ & $0.60 \pm 0.08$ & $0.60 \pm 0.05$ & {\color{green} $\mathbf{0.71 \pm 0.10}$} & nan  \\
1794324 & $0.52 \pm 0.04$ & {\color{green} $\mathbf{0.54 \pm 0.03}$} & $0.55 \pm 0.03$ & $0.57 \pm 0.01$ & {\color{green} $\mathbf{0.60 \pm 0.01}$} & {\color{green} $\mathbf{0.53 \pm 0.03}$} & $0.54 \pm 0.03$ & {\color{green} $\mathbf{0.56 \pm 0.03}$} & {\color{green} $\mathbf{0.58 \pm 0.02}$} & $0.60 \pm 0.02$ \\
1794504 & {\color{green} $\mathbf{0.72 \pm 0.06}$} & {\color{green} $\mathbf{0.75 \pm 0.05}$} & {\color{green} $\mathbf{0.81 \pm 0.04}$} & $0.85 \pm 0.24$ & nan & $0.67 \pm 0.04$ & $0.67 \pm 0.06$ & $0.73 \pm 0.05$ & {\color{green} $\mathbf{0.90 \pm 0.21}$} & nan  \\
1794519 & $0.69 \pm 0.08$ & {\color{green} $\mathbf{0.73 \pm 0.05}$} & $0.79 \pm 0.02$ & {\color{green} $\mathbf{0.81 \pm 0.05}$} & nan & {\color{green} $\mathbf{0.70 \pm 0.06}$} & $0.72 \pm 0.05$ & {\color{green} $\mathbf{0.80 \pm 0.03}$} & $0.77 \pm 0.07$ & nan  \\
1794557 & {\color{green} $\mathbf{0.52 \pm 0.04}$} & {\color{green} $\mathbf{0.53 \pm 0.03}$} & {\color{green} $\mathbf{0.56 \pm 0.04}$} & {\color{green} $\mathbf{0.58 \pm 0.04}$} & {\color{green} $\mathbf{0.60 \pm 0.02}$} & $0.51 \pm 0.03$ & $0.53 \pm 0.02$ & $0.53 \pm 0.03$ & $0.53 \pm 0.01$ & $0.54 \pm 0.03$ \\
1963701 & {\color{green} $\mathbf{0.54 \pm 0.03}$} & $0.54 \pm 0.02$ & {\color{green} $\mathbf{0.58 \pm 0.04}$} & {\color{green} $\mathbf{0.61 \pm 0.04}$} & {\color{green} $\mathbf{0.65 \pm 0.03}$} & $0.53 \pm 0.03$ & {\color{green} $\mathbf{0.54 \pm 0.04}$} & $0.56 \pm 0.03$ & $0.58 \pm 0.03$ & $0.63 \pm 0.03$ \\
1963705 & {\color{green} $\mathbf{0.67 \pm 0.09}$} & {\color{green} $\mathbf{0.75 \pm 0.05}$} & {\color{green} $\mathbf{0.78 \pm 0.03}$} & {\color{green} $\mathbf{0.82 \pm 0.03}$} & {\color{green} $\mathbf{0.84 \pm 0.02}$} & $0.65 \pm 0.07$ & $0.70 \pm 0.06$ & $0.73 \pm 0.04$ & $0.78 \pm 0.03$ & $0.82 \pm 0.02$ \\
1963715 & {\color{green} $\mathbf{0.61 \pm 0.05}$} & {\color{green} $\mathbf{0.59 \pm 0.08}$} & {\color{green} $\mathbf{0.68 \pm 0.04}$} & {\color{green} $\mathbf{0.72 \pm 0.04}$} & {\color{green} $\mathbf{0.75 \pm 0.02}$} & $0.50 \pm 0.06$ & $0.50 \pm 0.06$ & $0.56 \pm 0.04$ & $0.60 \pm 0.03$ & $0.65 \pm 0.03$ \\
1963721 & {\color{green} $\mathbf{0.45 \pm 0.08}$} & {\color{green} $\mathbf{0.48 \pm 0.08}$} & {\color{green} $\mathbf{0.55 \pm 0.06}$} & {\color{green} $\mathbf{0.57 \pm 0.07}$} & {\color{green} $\mathbf{0.71 \pm 0.09}$} & $0.38 \pm 0.05$ & $0.42 \pm 0.04$ & $0.44 \pm 0.03$ & $0.50 \pm 0.05$ & $0.61 \pm 0.08$ \\
1963723 & {\color{green} $\mathbf{0.70 \pm 0.10}$} & {\color{green} $\mathbf{0.75 \pm 0.09}$} & {\color{green} $\mathbf{0.80 \pm 0.03}$} & {\color{green} $\mathbf{0.84 \pm 0.02}$} & {\color{green} $\mathbf{0.86 \pm 0.02}$} & $0.61 \pm 0.05$ & $0.69 \pm 0.05$ & $0.74 \pm 0.04$ & $0.78 \pm 0.03$ & $0.81 \pm 0.02$ \\
1963731 & {\color{green} $\mathbf{0.74 \pm 0.06}$} & {\color{green} $\mathbf{0.79 \pm 0.03}$} & {\color{green} $\mathbf{0.82 \pm 0.02}$} & {\color{green} $\mathbf{0.87 \pm 0.02}$} & {\color{green} $\mathbf{0.89 \pm 0.02}$} & $0.63 \pm 0.08$ & $0.70 \pm 0.04$ & $0.75 \pm 0.02$ & $0.79 \pm 0.03$ & $0.82 \pm 0.02$ \\
1963741 & {\color{green} $\mathbf{0.60 \pm 0.05}$} & {\color{green} $\mathbf{0.65 \pm 0.05}$} & {\color{green} $\mathbf{0.70 \pm 0.04}$} & {\color{green} $\mathbf{0.75 \pm 0.03}$} & {\color{green} $\mathbf{0.78 \pm 0.02}$} & $0.51 \pm 0.08$ & $0.57 \pm 0.06$ & $0.59 \pm 0.05$ & $0.64 \pm 0.04$ & $0.71 \pm 0.03$ \\
1963756 & {\color{green} $\mathbf{0.63 \pm 0.09}$} & {\color{green} $\mathbf{0.75 \pm 0.04}$} & {\color{green} $\mathbf{0.77 \pm 0.03}$} & {\color{green} $\mathbf{0.80 \pm 0.02}$} & {\color{green} $\mathbf{0.83 \pm 0.02}$} & $0.57 \pm 0.07$ & $0.67 \pm 0.05$ & $0.69 \pm 0.06$ & $0.75 \pm 0.02$ & $0.79 \pm 0.02$ \\
1963773 & {\color{green} $\mathbf{0.59 \pm 0.06}$} & {\color{green} $\mathbf{0.61 \pm 0.07}$} & {\color{green} $\mathbf{0.65 \pm 0.04}$} & {\color{green} $\mathbf{0.69 \pm 0.03}$} & {\color{green} $\mathbf{0.72 \pm 0.02}$} & $0.54 \pm 0.06$ & $0.57 \pm 0.06$ & $0.61 \pm 0.05$ & $0.65 \pm 0.03$ & $0.70 \pm 0.03$ \\
1963799 & {\color{green} $\mathbf{0.58 \pm 0.07}$} & {\color{green} $\mathbf{0.64 \pm 0.04}$} & {\color{green} $\mathbf{0.68 \pm 0.04}$} & {\color{green} $\mathbf{0.71 \pm 0.03}$} & {\color{green} $\mathbf{0.74 \pm 0.04}$} & $0.48 \pm 0.06$ & $0.55 \pm 0.05$ & $0.58 \pm 0.07$ & $0.63 \pm 0.05$ & $0.68 \pm 0.07$ \\
1963810 & {\color{green} $\mathbf{0.78 \pm 0.06}$} & {\color{green} $\mathbf{0.81 \pm 0.05}$} & {\color{green} $\mathbf{0.85 \pm 0.02}$} & {\color{green} $\mathbf{0.86 \pm 0.02}$} & {\color{green} $\mathbf{0.88 \pm 0.01}$} & $0.67 \pm 0.06$ & $0.73 \pm 0.05$ & $0.76 \pm 0.06$ & $0.80 \pm 0.04$ & $0.84 \pm 0.02$ \\
1963818 & {\color{green} $\mathbf{0.71 \pm 0.05}$} & {\color{green} $\mathbf{0.74 \pm 0.04}$} & {\color{green} $\mathbf{0.78 \pm 0.03}$} & {\color{green} $\mathbf{0.80 \pm 0.03}$} & {\color{green} $\mathbf{0.82 \pm 0.02}$} & $0.66 \pm 0.07$ & $0.68 \pm 0.06$ & $0.72 \pm 0.03$ & $0.76 \pm 0.03$ & $0.79 \pm 0.02$ \\
1963819 & {\color{green} $\mathbf{0.61 \pm 0.07}$} & {\color{green} $\mathbf{0.67 \pm 0.06}$} & {\color{green} $\mathbf{0.73 \pm 0.06}$} & {\color{green} $\mathbf{0.77 \pm 0.05}$} & {\color{green} $\mathbf{0.82 \pm 0.01}$} & $0.48 \pm 0.05$ & $0.56 \pm 0.07$ & $0.60 \pm 0.08$ & $0.68 \pm 0.06$ & $0.77 \pm 0.02$ \\
1963824 & {\color{green} $\mathbf{0.59 \pm 0.05}$} & {\color{green} $\mathbf{0.66 \pm 0.05}$} & {\color{green} $\mathbf{0.70 \pm 0.03}$} & {\color{green} $\mathbf{0.74 \pm 0.03}$} & {\color{green} $\mathbf{0.77 \pm 0.05}$} & $0.54 \pm 0.04$ & $0.59 \pm 0.05$ & $0.64 \pm 0.04$ & $0.64 \pm 0.04$ & $0.69 \pm 0.03$ \\
1963825 & {\color{green} $\mathbf{0.62 \pm 0.06}$} & {\color{green} $\mathbf{0.69 \pm 0.04}$} & {\color{green} $\mathbf{0.70 \pm 0.05}$} & {\color{green} $\mathbf{0.75 \pm 0.05}$} & {\color{green} $\mathbf{0.81 \pm 0.04}$} & $0.57 \pm 0.04$ & $0.61 \pm 0.04$ & $0.64 \pm 0.04$ & $0.65 \pm 0.04$ & $0.69 \pm 0.03$ \\
1963827 & {\color{green} $\mathbf{0.74 \pm 0.09}$} & {\color{green} $\mathbf{0.78 \pm 0.06}$} & {\color{green} $\mathbf{0.84 \pm 0.03}$} & {\color{green} $\mathbf{0.86 \pm 0.01}$} & {\color{green} $\mathbf{0.88 \pm 0.02}$} & $0.71 \pm 0.06$ & $0.72 \pm 0.08$ & $0.79 \pm 0.05$ & $0.82 \pm 0.02$ & $0.84 \pm 0.02$ \\
1963831 & {\color{green} $\mathbf{0.61 \pm 0.10}$} & {\color{green} $\mathbf{0.69 \pm 0.07}$} & {\color{green} $\mathbf{0.76 \pm 0.02}$} & {\color{green} $\mathbf{0.78 \pm 0.04}$} & {\color{green} $\mathbf{0.81 \pm 0.05}$} & $0.55 \pm 0.11$ & $0.61 \pm 0.07$ & $0.69 \pm 0.06$ & $0.74 \pm 0.05$ & $0.77 \pm 0.04$ \\
1963838 & {\color{green} $\mathbf{0.56 \pm 0.04}$} & {\color{green} $\mathbf{0.57 \pm 0.05}$} & {\color{green} $\mathbf{0.59 \pm 0.06}$} & {\color{green} $\mathbf{0.67 \pm 0.05}$} & nan & $0.54 \pm 0.05$ & $0.55 \pm 0.04$ & $0.58 \pm 0.04$ & $0.62 \pm 0.06$ & nan  \\
        \bottomrule
    \end{tabular}
}
\vspace{-0.2cm}
\label{table:50_mt_fsmol}
\end{table*}

\begin{table*}[h!]
\centering
\caption{PROTO Results measuring $\Delta$AUPRC on the first 50 tasks in the test set of FS-Mol.}
\resizebox{1.0\linewidth}{!}{%
    \begin{tabular}{lllllllllll}
    \toprule
        {TASK-ID} & {16 (\method-PROTO)} & {32 (\method-PROTO)} & {64 (\method-PROTO)} & {128 (\method-PROTO)} & {256 (\method-PROTO)} & {16 (PROTO)} & {32 (PROTO)} & {64 (PROTO)} & {128 (PROTO)} & {256 (PROTO)} \\
    \midrule
    1006005 & {\color{green} $\mathbf{0.61 \pm 0.05}$} & {\color{green} $\mathbf{0.63 \pm 0.04}$} & {\color{green} $\mathbf{0.66 \pm 0.04}$} & {\color{green} $\mathbf{0.69 \pm 0.06}$} & nan & $0.56 \pm 0.06$ & $0.61 \pm 0.05$ & $0.65 \pm 0.03$ & $0.66 \pm 0.07$ & nan  \\
    1066254 & {\color{green} $\mathbf{0.72 \pm 0.07}$} & $0.77 \pm 0.06$ & {\color{green} $\mathbf{0.84 \pm 0.04}$} & {\color{green} $\mathbf{0.87 \pm 0.10}$} & nan & $0.71 \pm 0.08$ & {\color{green} $\mathbf{0.78 \pm 0.06}$} & $0.82 \pm 0.06$ & $0.85 \pm 0.08$ & nan  \\
    1119333 & {\color{green} $\mathbf{0.73 \pm 0.05}$} & {\color{green} $\mathbf{0.74 \pm 0.06}$} & {\color{green} $\mathbf{0.78 \pm 0.03}$} & {\color{green} $\mathbf{0.82 \pm 0.02}$} & {\color{green} $\mathbf{0.84 \pm 0.04}$} & $0.71 \pm 0.06$ & $0.72 \pm 0.04$ & {\color{green} $\mathbf{0.78 \pm 0.03}$} & $0.81 \pm 0.02$ & $0.83 \pm 0.04$ \\
    1243967 & {\color{green} $\mathbf{0.78 \pm 0.03}$} & {\color{green} $\mathbf{0.81 \pm 0.04}$} & {\color{green} $\mathbf{0.83 \pm 0.03}$} & {\color{green} $\mathbf{0.84 \pm 0.03}$} & nan & $0.73 \pm 0.05$ & $0.76 \pm 0.04$ & $0.80 \pm 0.03$ & $0.81 \pm 0.03$ & nan  \\
    1243970 & {\color{green} $\mathbf{0.75 \pm 0.05}$} & {\color{green} $\mathbf{0.79 \pm 0.04}$} & {\color{green} $\mathbf{0.81 \pm 0.03}$} & {\color{green} $\mathbf{0.85 \pm 0.03}$} & nan & $0.71 \pm 0.07$ & $0.74 \pm 0.04$ & $0.81 \pm 0.03$ & $0.82 \pm 0.05$ & nan  \\
    1613777 & $0.52 \pm 0.02$ & {\color{green} $\mathbf{0.54 \pm 0.03}$} & $0.56 \pm 0.03$ & $0.59 \pm 0.02$ & $0.61 \pm 0.02$ & {\color{green} $\mathbf{0.52 \pm 0.04}$} & $0.54 \pm 0.04$ & {\color{green} $\mathbf{0.57 \pm 0.03}$} & {\color{green} $\mathbf{0.60 \pm 0.04}$} & {\color{green} $\mathbf{0.62 \pm 0.03}$} \\
    1613800 & {\color{green} $\mathbf{0.42 \pm 0.02}$} & {\color{green} $\mathbf{0.44 \pm 0.02}$} & {\color{green} $\mathbf{0.46 \pm 0.02}$} & {\color{green} $\mathbf{0.48 \pm 0.02}$} & {\color{green} $\mathbf{0.50 \pm 0.01}$} & $0.41 \pm 0.03$ & $0.43 \pm 0.03$ & $0.45 \pm 0.02$ & $0.46 \pm 0.02$ & $0.48 \pm 0.02$ \\
    1613898 & $0.52 \pm 0.02$ & $0.55 \pm 0.04$ & $0.54 \pm 0.04$ & {\color{green} $\mathbf{0.61 \pm 0.07}$} & nan & {\color{green} $\mathbf{0.55 \pm 0.04}$} & {\color{green} $\mathbf{0.55 \pm 0.05}$} & {\color{green} $\mathbf{0.55 \pm 0.03}$} & $0.57 \pm 0.08$ & nan  \\
    1613907 & {\color{green} $\mathbf{0.56 \pm 0.04}$} & $0.58 \pm 0.04$ & $0.63 \pm 0.04$ & $0.64 \pm 0.14$ & nan & $0.55 \pm 0.06$ & {\color{green} $\mathbf{0.62 \pm 0.06}$} & {\color{green} $\mathbf{0.66 \pm 0.07}$} & {\color{green} $\mathbf{0.68 \pm 0.10}$} & nan  \\
    1613926 & {\color{green} $\mathbf{0.70 \pm 0.04}$} & {\color{green} $\mathbf{0.74 \pm 0.06}$} & {\color{green} $\mathbf{0.77 \pm 0.04}$} & {\color{green} $\mathbf{0.88 \pm 0.12}$} & nan & $0.65 \pm 0.08$ & $0.70 \pm 0.06$ & $0.76 \pm 0.04$ & $0.86 \pm 0.15$ & nan  \\
    1613949 & {\color{green} $\mathbf{0.57 \pm 0.08}$} & {\color{green} $\mathbf{0.63 \pm 0.05}$} & {\color{green} $\mathbf{0.67 \pm 0.06}$} & {\color{green} $\mathbf{0.67 \pm 0.10}$} & nan & $0.53 \pm 0.05$ & $0.57 \pm 0.04$ & $0.58 \pm 0.05$ & $0.63 \pm 0.10$ & nan  \\
    1614027 & $0.53 \pm 0.02$ & {\color{green} $\mathbf{0.57 \pm 0.02}$} & $0.60 \pm 0.03$ & $0.64 \pm 0.02$ & {\color{green} $\mathbf{0.67 \pm 0.02}$} & {\color{green} $\mathbf{0.55 \pm 0.02}$} & $0.57 \pm 0.04$ & {\color{green} $\mathbf{0.61 \pm 0.04}$} & {\color{green} $\mathbf{0.64 \pm 0.02}$} & $0.67 \pm 0.02$ \\
    1614153 & {\color{green} $\mathbf{0.35 \pm 0.02}$} & $0.37 \pm 0.01$ & $0.36 \pm 0.02$ & $0.39 \pm 0.01$ & {\color{green} $\mathbf{0.41 \pm 0.01}$} & $0.34 \pm 0.03$ & {\color{green} $\mathbf{0.37 \pm 0.02}$} & {\color{green} $\mathbf{0.37 \pm 0.02}$} & {\color{green} $\mathbf{0.39 \pm 0.01}$} & {\color{green} $\mathbf{0.41 \pm 0.01}$} \\
    1614292 & {\color{green} $\mathbf{0.36 \pm 0.02}$} & $0.38 \pm 0.03$ & $0.39 \pm 0.02$ & $0.40 \pm 0.02$ & $0.41 \pm 0.02$ & {\color{green} $\mathbf{0.36 \pm 0.02}$} & {\color{green} $\mathbf{0.39 \pm 0.02}$} & {\color{green} $\mathbf{0.41 \pm 0.02}$} & {\color{green} $\mathbf{0.42 \pm 0.03}$} & {\color{green} $\mathbf{0.44 \pm 0.02}$} \\
    1614423 & {\color{green} $\mathbf{0.68 \pm 0.06}$} & {\color{green} $\mathbf{0.72 \pm 0.04}$} & {\color{green} $\mathbf{0.76 \pm 0.02}$} & {\color{green} $\mathbf{0.78 \pm 0.02}$} & {\color{green} $\mathbf{0.79 \pm 0.04}$} & $0.54 \pm 0.07$ & $0.58 \pm 0.05$ & $0.67 \pm 0.04$ & $0.73 \pm 0.02$ & $0.77 \pm 0.02$ \\
    1614433 & {\color{green} $\mathbf{0.49 \pm 0.04}$} & {\color{green} $\mathbf{0.50 \pm 0.04}$} & {\color{green} $\mathbf{0.51 \pm 0.03}$} & {\color{green} $\mathbf{0.53 \pm 0.02}$} & $0.54 \pm 0.02$ & $0.46 \pm 0.04$ & $0.48 \pm 0.02$ & $0.48 \pm 0.03$ & $0.52 \pm 0.03$ & {\color{green} $\mathbf{0.54 \pm 0.04}$} \\
    1614466 & $0.47 \pm 0.03$ & $0.48 \pm 0.04$ & $0.49 \pm 0.01$ & $0.51 \pm 0.03$ & $0.53 \pm 0.03$ & {\color{green} $\mathbf{0.48 \pm 0.05}$} & {\color{green} $\mathbf{0.53 \pm 0.05}$} & {\color{green} $\mathbf{0.53 \pm 0.04}$} & {\color{green} $\mathbf{0.57 \pm 0.02}$} & {\color{green} $\mathbf{0.58 \pm 0.01}$} \\
    1614503 & {\color{green} $\mathbf{0.49 \pm 0.03}$} & $0.51 \pm 0.07$ & $0.56 \pm 0.06$ & $0.70 \pm 0.22$ & nan & $0.49 \pm 0.07$ & {\color{green} $\mathbf{0.55 \pm 0.05}$} & {\color{green} $\mathbf{0.59 \pm 0.02}$} & {\color{green} $\mathbf{0.76 \pm 0.22}$} & nan  \\
    1614508 & $0.84 \pm 0.02$ & $0.85 \pm 0.03$ & $0.86 \pm 0.03$ & {\color{green} $\mathbf{0.89 \pm 0.04}$} & nan & {\color{green} $\mathbf{0.86 \pm 0.02}$} & {\color{green} $\mathbf{0.86 \pm 0.02}$} & {\color{green} $\mathbf{0.87 \pm 0.02}$} & $0.88 \pm 0.04$ & nan  \\
    1614522 & {\color{green} $\mathbf{0.58 \pm 0.04}$} & {\color{green} $\mathbf{0.61 \pm 0.03}$} & $0.60 \pm 0.03$ & {\color{green} $\mathbf{0.64 \pm 0.02}$} & $0.67 \pm 0.03$ & {\color{green} $\mathbf{0.58 \pm 0.03}$} & $0.60 \pm 0.03$ & {\color{green} $\mathbf{0.61 \pm 0.03}$} & {\color{green} $\mathbf{0.64 \pm 0.02}$} & {\color{green} $\mathbf{0.67 \pm 0.02}$} \\
    1737951 & {\color{green} $\mathbf{0.65 \pm 0.08}$} & {\color{green} $\mathbf{0.71 \pm 0.06}$} & {\color{green} $\mathbf{0.76 \pm 0.04}$} & {\color{green} $\mathbf{0.80 \pm 0.06}$} & nan & $0.63 \pm 0.06$ & $0.67 \pm 0.05$ & $0.71 \pm 0.04$ & $0.75 \pm 0.07$ & nan  \\
    1738079 & $0.49 \pm 0.03$ & $0.49 \pm 0.03$ & $0.50 \pm 0.04$ & $0.49 \pm 0.04$ & nan & {\color{green} $\mathbf{0.49 \pm 0.02}$} & {\color{green} $\mathbf{0.49 \pm 0.03}$} & {\color{green} $\mathbf{0.50 \pm 0.03}$} & {\color{green} $\mathbf{0.54 \pm 0.06}$} & nan  \\
    1738362 & $0.46 \pm 0.06$ & $0.55 \pm 0.08$ & $0.59 \pm 0.06$ & $0.64 \pm 0.22$ & nan & {\color{green} $\mathbf{0.49 \pm 0.10}$} & {\color{green} $\mathbf{0.56 \pm 0.09}$} & {\color{green} $\mathbf{0.65 \pm 0.05}$} & {\color{green} $\mathbf{0.82 \pm 0.17}$} & nan  \\
    1738395 & $0.53 \pm 0.05$ & $0.51 \pm 0.04$ & $0.52 \pm 0.03$ & $0.57 \pm 0.04$ & nan & {\color{green} $\mathbf{0.53 \pm 0.04}$} & {\color{green} $\mathbf{0.53 \pm 0.04}$} & {\color{green} $\mathbf{0.54 \pm 0.05}$} & {\color{green} $\mathbf{0.58 \pm 0.06}$} & nan  \\
    1738485 & $0.54 \pm 0.04$ & $0.57 \pm 0.04$ & $0.58 \pm 0.05$ & {\color{green} $\mathbf{0.63 \pm 0.04}$} & {\color{green} $\mathbf{0.69 \pm 0.05}$} & {\color{green} $\mathbf{0.58 \pm 0.03}$} & {\color{green} $\mathbf{0.59 \pm 0.03}$} & {\color{green} $\mathbf{0.61 \pm 0.04}$} & $0.62 \pm 0.03$ & $0.64 \pm 0.07$ \\
    1738502 & $0.41 \pm 0.07$ & $0.44 \pm 0.05$ & $0.47 \pm 0.04$ & $0.49 \pm 0.03$ & {\color{green} $\mathbf{0.53 \pm 0.03}$} & {\color{green} $\mathbf{0.42 \pm 0.07}$} & {\color{green} $\mathbf{0.45 \pm 0.05}$} & {\color{green} $\mathbf{0.48 \pm 0.03}$} & {\color{green} $\mathbf{0.49 \pm 0.02}$} & $0.52 \pm 0.02$ \\
    1738573 & $0.52 \pm 0.03$ & {\color{green} $\mathbf{0.53 \pm 0.03}$} & {\color{green} $\mathbf{0.54 \pm 0.02}$} & {\color{green} $\mathbf{0.56 \pm 0.02}$} & {\color{green} $\mathbf{0.57 \pm 0.02}$} & {\color{green} $\mathbf{0.53 \pm 0.02}$} & $0.52 \pm 0.02$ & $0.53 \pm 0.02$ & $0.55 \pm 0.01$ & $0.55 \pm 0.01$ \\
    1738579 & $0.52 \pm 0.04$ & $0.55 \pm 0.05$ & $0.58 \pm 0.02$ & {\color{green} $\mathbf{0.65 \pm 0.06}$} & nan & {\color{green} $\mathbf{0.55 \pm 0.06}$} & {\color{green} $\mathbf{0.59 \pm 0.03}$} & {\color{green} $\mathbf{0.59 \pm 0.03}$} & $0.64 \pm 0.04$ & nan  \\
    1738633 & $0.56 \pm 0.04$ & $0.58 \pm 0.04$ & {\color{green} $\mathbf{0.59 \pm 0.05}$} & {\color{green} $\mathbf{0.63 \pm 0.08}$} & nan & {\color{green} $\mathbf{0.56 \pm 0.05}$} & {\color{green} $\mathbf{0.59 \pm 0.05}$} & $0.59 \pm 0.04$ & $0.62 \pm 0.13$ & nan  \\
    1794324 & $0.50 \pm 0.03$ & $0.51 \pm 0.03$ & $0.53 \pm 0.03$ & $0.57 \pm 0.02$ & $0.60 \pm 0.02$ & {\color{green} $\mathbf{0.53 \pm 0.03}$} & {\color{green} $\mathbf{0.54 \pm 0.02}$} & {\color{green} $\mathbf{0.56 \pm 0.03}$} & {\color{green} $\mathbf{0.59 \pm 0.02}$} & {\color{green} $\mathbf{0.62 \pm 0.02}$} \\
    1794504 & {\color{green} $\mathbf{0.70 \pm 0.06}$} & {\color{green} $\mathbf{0.72 \pm 0.05}$} & {\color{green} $\mathbf{0.80 \pm 0.06}$} & {\color{green} $\mathbf{0.95 \pm 0.16}$} & nan & $0.59 \pm 0.05$ & $0.64 \pm 0.05$ & $0.72 \pm 0.09$ & {\color{green} $\mathbf{0.95 \pm 0.16}$} & nan  \\
    1794519 & $0.65 \pm 0.08$ & $0.72 \pm 0.06$ & $0.78 \pm 0.04$ & {\color{green} $\mathbf{0.82 \pm 0.05}$} & nan & {\color{green} $\mathbf{0.73 \pm 0.06}$} & {\color{green} $\mathbf{0.78 \pm 0.03}$} & {\color{green} $\mathbf{0.81 \pm 0.02}$} & $0.79 \pm 0.04$ & nan  \\
    1794557 & {\color{green} $\mathbf{0.54 \pm 0.02}$} & {\color{green} $\mathbf{0.54 \pm 0.02}$} & {\color{green} $\mathbf{0.55 \pm 0.02}$} & {\color{green} $\mathbf{0.57 \pm 0.03}$} & {\color{green} $\mathbf{0.60 \pm 0.02}$} & $0.52 \pm 0.02$ & $0.53 \pm 0.02$ & $0.53 \pm 0.03$ & $0.54 \pm 0.03$ & $0.56 \pm 0.02$ \\
    1963701 & {\color{green} $\mathbf{0.58 \pm 0.05}$} & {\color{green} $\mathbf{0.59 \pm 0.06}$} & {\color{green} $\mathbf{0.62 \pm 0.03}$} & {\color{green} $\mathbf{0.67 \pm 0.02}$} & {\color{green} $\mathbf{0.69 \pm 0.02}$} & $0.56 \pm 0.04$ & $0.58 \pm 0.05$ & $0.60 \pm 0.04$ & $0.62 \pm 0.03$ & $0.66 \pm 0.02$ \\
    1963705 & {\color{green} $\mathbf{0.74 \pm 0.06}$} & {\color{green} $\mathbf{0.77 \pm 0.04}$} & {\color{green} $\mathbf{0.81 \pm 0.02}$} & {\color{green} $\mathbf{0.83 \pm 0.01}$} & {\color{green} $\mathbf{0.84 \pm 0.01}$} & $0.71 \pm 0.05$ & $0.74 \pm 0.03$ & $0.78 \pm 0.03$ & $0.80 \pm 0.02$ & $0.81 \pm 0.02$ \\
    1963715 & {\color{green} $\mathbf{0.63 \pm 0.10}$} & {\color{green} $\mathbf{0.68 \pm 0.05}$} & {\color{green} $\mathbf{0.72 \pm 0.03}$} & {\color{green} $\mathbf{0.76 \pm 0.02}$} & {\color{green} $\mathbf{0.78 \pm 0.02}$} & $0.57 \pm 0.07$ & $0.61 \pm 0.03$ & $0.65 \pm 0.03$ & $0.69 \pm 0.02$ & $0.72 \pm 0.02$ \\
    1963721 & {\color{green} $\mathbf{0.47 \pm 0.07}$} & {\color{green} $\mathbf{0.52 \pm 0.04}$} & {\color{green} $\mathbf{0.54 \pm 0.04}$} & {\color{green} $\mathbf{0.58 \pm 0.04}$} & {\color{green} $\mathbf{0.63 \pm 0.08}$} & $0.45 \pm 0.06$ & $0.47 \pm 0.04$ & $0.51 \pm 0.04$ & $0.53 \pm 0.03$ & $0.61 \pm 0.10$ \\
    1963723 & {\color{green} $\mathbf{0.77 \pm 0.05}$} & {\color{green} $\mathbf{0.80 \pm 0.04}$} & {\color{green} $\mathbf{0.84 \pm 0.01}$} & {\color{green} $\mathbf{0.86 \pm 0.01}$} & {\color{green} $\mathbf{0.87 \pm 0.01}$} & $0.72 \pm 0.08$ & $0.79 \pm 0.03$ & $0.81 \pm 0.01$ & $0.83 \pm 0.01$ & $0.85 \pm 0.01$ \\
    1963731 & $0.79 \pm 0.06$ & {\color{green} $\mathbf{0.84 \pm 0.02}$} & {\color{green} $\mathbf{0.87 \pm 0.02}$} & {\color{green} $\mathbf{0.89 \pm 0.01}$} & {\color{green} $\mathbf{0.90 \pm 0.01}$} & {\color{green} $\mathbf{0.80 \pm 0.05}$} & $0.83 \pm 0.02$ & $0.86 \pm 0.01$ & $0.87 \pm 0.01$ & $0.88 \pm 0.01$ \\
    1963741 & {\color{green} $\mathbf{0.64 \pm 0.06}$} & {\color{green} $\mathbf{0.70 \pm 0.04}$} & {\color{green} $\mathbf{0.74 \pm 0.02}$} & {\color{green} $\mathbf{0.76 \pm 0.02}$} & {\color{green} $\mathbf{0.78 \pm 0.02}$} & $0.59 \pm 0.03$ & $0.63 \pm 0.04$ & $0.68 \pm 0.04$ & $0.71 \pm 0.02$ & $0.73 \pm 0.02$ \\
    1963756 & {\color{green} $\mathbf{0.74 \pm 0.06}$} & {\color{green} $\mathbf{0.77 \pm 0.03}$} & {\color{green} $\mathbf{0.79 \pm 0.02}$} & {\color{green} $\mathbf{0.82 \pm 0.01}$} & {\color{green} $\mathbf{0.84 \pm 0.01}$} & $0.63 \pm 0.06$ & $0.68 \pm 0.04$ & $0.73 \pm 0.02$ & $0.76 \pm 0.02$ & $0.77 \pm 0.03$ \\
    1963773 & {\color{green} $\mathbf{0.62 \pm 0.09}$} & $0.64 \pm 0.06$ & {\color{green} $\mathbf{0.69 \pm 0.04}$} & $0.72 \pm 0.03$ & {\color{green} $\mathbf{0.75 \pm 0.01}$} & $0.61 \pm 0.07$ & {\color{green} $\mathbf{0.65 \pm 0.06}$} & $0.69 \pm 0.04$ & {\color{green} $\mathbf{0.73 \pm 0.02}$} & $0.74 \pm 0.03$ \\
    1963799 & {\color{green} $\mathbf{0.65 \pm 0.07}$} & {\color{green} $\mathbf{0.67 \pm 0.05}$} & {\color{green} $\mathbf{0.69 \pm 0.02}$} & {\color{green} $\mathbf{0.74 \pm 0.02}$} & {\color{green} $\mathbf{0.76 \pm 0.03}$} & $0.56 \pm 0.07$ & $0.63 \pm 0.04$ & $0.67 \pm 0.03$ & $0.69 \pm 0.02$ & $0.72 \pm 0.05$ \\
    1963810 & {\color{green} $\mathbf{0.80 \pm 0.04}$} & {\color{green} $\mathbf{0.83 \pm 0.03}$} & {\color{green} $\mathbf{0.86 \pm 0.01}$} & {\color{green} $\mathbf{0.86 \pm 0.01}$} & {\color{green} $\mathbf{0.88 \pm 0.01}$} & $0.76 \pm 0.06$ & $0.79 \pm 0.04$ & $0.82 \pm 0.02$ & $0.85 \pm 0.01$ & $0.87 \pm 0.01$ \\
    1963818 & {\color{green} $\mathbf{0.71 \pm 0.06}$} & {\color{green} $\mathbf{0.75 \pm 0.05}$} & $0.78 \pm 0.03$ & {\color{green} $\mathbf{0.82 \pm 0.02}$} & {\color{green} $\mathbf{0.84 \pm 0.02}$} & {\color{green} $\mathbf{0.71 \pm 0.03}$} & $0.73 \pm 0.04$ & {\color{green} $\mathbf{0.78 \pm 0.02}$} & $0.81 \pm 0.01$ & $0.82 \pm 0.01$ \\
    1963819 & {\color{green} $\mathbf{0.65 \pm 0.08}$} & $0.68 \pm 0.06$ & {\color{green} $\mathbf{0.74 \pm 0.03}$} & $0.78 \pm 0.03$ & $0.81 \pm 0.02$ & $0.62 \pm 0.12$ & {\color{green} $\mathbf{0.70 \pm 0.05}$} & $0.73 \pm 0.05$ & {\color{green} $\mathbf{0.79 \pm 0.02}$} & {\color{green} $\mathbf{0.81 \pm 0.02}$} \\
    1963824 & {\color{green} $\mathbf{0.60 \pm 0.06}$} & {\color{green} $\mathbf{0.63 \pm 0.04}$} & {\color{green} $\mathbf{0.67 \pm 0.05}$} & {\color{green} $\mathbf{0.72 \pm 0.04}$} & {\color{green} $\mathbf{0.75 \pm 0.03}$} & $0.52 \pm 0.04$ & $0.54 \pm 0.03$ & $0.57 \pm 0.03$ & $0.61 \pm 0.03$ & $0.62 \pm 0.06$ \\
    1963825 & {\color{green} $\mathbf{0.66 \pm 0.06}$} & {\color{green} $\mathbf{0.70 \pm 0.05}$} & {\color{green} $\mathbf{0.72 \pm 0.03}$} & {\color{green} $\mathbf{0.74 \pm 0.02}$} & {\color{green} $\mathbf{0.78 \pm 0.04}$} & $0.62 \pm 0.07$ & $0.68 \pm 0.03$ & $0.71 \pm 0.04$ & $0.73 \pm 0.02$ & $0.77 \pm 0.03$ \\
    1963827 & {\color{green} $\mathbf{0.78 \pm 0.06}$} & {\color{green} $\mathbf{0.81 \pm 0.03}$} & {\color{green} $\mathbf{0.86 \pm 0.02}$} & {\color{green} $\mathbf{0.88 \pm 0.00}$} & {\color{green} $\mathbf{0.88 \pm 0.01}$} & $0.77 \pm 0.04$ & $0.80 \pm 0.04$ & $0.84 \pm 0.02$ & $0.86 \pm 0.01$ & $0.87 \pm 0.01$ \\
    1963831 & {\color{green} $\mathbf{0.62 \pm 0.10}$} & $0.69 \pm 0.07$ & $0.74 \pm 0.03$ & $0.77 \pm 0.03$ & $0.78 \pm 0.05$ & $0.61 \pm 0.08$ & {\color{green} $\mathbf{0.69 \pm 0.05}$} & {\color{green} $\mathbf{0.76 \pm 0.03}$} & {\color{green} $\mathbf{0.78 \pm 0.02}$} & {\color{green} $\mathbf{0.79 \pm 0.06}$} \\
    1963838 & $0.52 \pm 0.03$ & $0.55 \pm 0.02$ & $0.58 \pm 0.03$ & $0.60 \pm 0.03$ & nan & {\color{green} $\mathbf{0.55 \pm 0.03}$} & {\color{green} $\mathbf{0.57 \pm 0.05}$} & {\color{green} $\mathbf{0.61 \pm 0.04}$} & {\color{green} $\mathbf{0.64 \pm 0.03}$} & nan  \\
    \bottomrule
    \end{tabular}
}
\vspace{-0.2cm}
\label{table:50_proto_fsmol}
\end{table*}

\begin{table*}[h!]
\centering
\caption{MAML results measuring $\Delta$AUPRC on the first 50 tasks in the test set of FS-Mol.}
\resizebox{1.0\linewidth}{!}{%
    \begin{tabular}{lllllllllll}
    \toprule
        {TASK-ID} & {16 (\method-MAML)} & {32 (\method-MAML)} & {64 (\method-MAML)} & {128 (\method-MAML)} & {256 (\method-MAML)} & {16 (MAML)} & {32 (MAML)} & {64 (MAML)} & {128 (MAML)} & {256 (MAML)} \\
    \midrule
    1006005 & $0.47 \pm 0.03$ & $0.49 \pm 0.05$ & $0.49 \pm 0.04$ & $0.54 \pm 0.06$ & nan & {\color{green} $\mathbf{0.49 \pm 0.02}$} & {\color{green} $\mathbf{0.51 \pm 0.04}$} & {\color{green} $\mathbf{0.52 \pm 0.03}$} & {\color{green} $\mathbf{0.56 \pm 0.05}$} & nan  \\
    1066254 & {\color{green} $\mathbf{0.58 \pm 0.07}$} & {\color{green} $\mathbf{0.61 \pm 0.05}$} & {\color{green} $\mathbf{0.65 \pm 0.07}$} & {\color{green} $\mathbf{0.65 \pm 0.13}$} & nan & $0.56 \pm 0.07$ & $0.58 \pm 0.04$ & $0.59 \pm 0.07$ & $0.63 \pm 0.15$ & nan  \\
    1119333 & {\color{green} $\mathbf{0.73 \pm 0.03}$} & {\color{green} $\mathbf{0.74 \pm 0.03}$} & {\color{green} $\mathbf{0.77 \pm 0.04}$} & {\color{green} $\mathbf{0.76 \pm 0.05}$} & {\color{green} $\mathbf{0.77 \pm 0.06}$} & $0.70 \pm 0.03$ & $0.72 \pm 0.04$ & $0.74 \pm 0.05$ & $0.72 \pm 0.02$ & $0.74 \pm 0.05$ \\
    1243967 & $0.72 \pm 0.03$ & {\color{green} $\mathbf{0.76 \pm 0.03}$} & {\color{green} $\mathbf{0.75 \pm 0.04}$} & {\color{green} $\mathbf{0.77 \pm 0.04}$} & nan & {\color{green} $\mathbf{0.72 \pm 0.05}$} & $0.75 \pm 0.01$ & $0.74 \pm 0.04$ & $0.76 \pm 0.04$ & nan  \\
    1243970 & $0.70 \pm 0.04$ & $0.71 \pm 0.01$ & $0.70 \pm 0.02$ & {\color{green} $\mathbf{0.71 \pm 0.03}$} & nan & {\color{green} $\mathbf{0.71 \pm 0.03}$} & {\color{green} $\mathbf{0.72 \pm 0.03}$} & {\color{green} $\mathbf{0.72 \pm 0.03}$} & $0.68 \pm 0.03$ & nan  \\
    1613777 & {\color{green} $\mathbf{0.53 \pm 0.01}$} & {\color{green} $\mathbf{0.53 \pm 0.01}$} & {\color{green} $\mathbf{0.53 \pm 0.01}$} & {\color{green} $\mathbf{0.54 \pm 0.02}$} & {\color{green} $\mathbf{0.54 \pm 0.02}$} & $0.50 \pm 0.01$ & $0.51 \pm 0.01$ & $0.51 \pm 0.01$ & $0.53 \pm 0.02$ & $0.53 \pm 0.02$ \\
    1613800 & {\color{green} $\mathbf{0.40 \pm 0.00}$} & {\color{green} $\mathbf{0.40 \pm 0.01}$} & {\color{green} $\mathbf{0.41 \pm 0.01}$} & {\color{green} $\mathbf{0.42 \pm 0.01}$} & {\color{green} $\mathbf{0.42 \pm 0.01}$} & {\color{green} $\mathbf{0.40 \pm 0.01}$} & $0.40 \pm 0.01$ & $0.40 \pm 0.01$ & $0.41 \pm 0.01$ & $0.41 \pm 0.01$ \\
    1613898 & $0.56 \pm 0.02$ & $0.55 \pm 0.04$ & {\color{green} $\mathbf{0.58 \pm 0.04}$} & {\color{green} $\mathbf{0.57 \pm 0.08}$} & nan & {\color{green} $\mathbf{0.57 \pm 0.02}$} & {\color{green} $\mathbf{0.56 \pm 0.03}$} & $0.56 \pm 0.03$ & $0.56 \pm 0.04$ & nan  \\
    1613907 & {\color{green} $\mathbf{0.54 \pm 0.03}$} & {\color{green} $\mathbf{0.55 \pm 0.04}$} & {\color{green} $\mathbf{0.54 \pm 0.06}$} & {\color{green} $\mathbf{0.55 \pm 0.07}$} & nan & $0.53 \pm 0.03$ & $0.51 \pm 0.02$ & $0.53 \pm 0.05$ & $0.55 \pm 0.13$ & nan  \\
    1613926 & {\color{green} $\mathbf{0.67 \pm 0.04}$} & {\color{green} $\mathbf{0.65 \pm 0.06}$} & {\color{green} $\mathbf{0.71 \pm 0.06}$} & {\color{green} $\mathbf{0.78 \pm 0.20}$} & nan & $0.57 \pm 0.09$ & $0.59 \pm 0.10$ & $0.65 \pm 0.12$ & $0.77 \pm 0.12$ & nan  \\
    1613949 & $0.46 \pm 0.04$ & $0.47 \pm 0.04$ & $0.47 \pm 0.05$ & {\color{green} $\mathbf{0.45 \pm 0.10}$} & nan & {\color{green} $\mathbf{0.48 \pm 0.03}$} & {\color{green} $\mathbf{0.48 \pm 0.04}$} & {\color{green} $\mathbf{0.48 \pm 0.06}$} & $0.43 \pm 0.09$ & nan  \\
    1614027 & $0.51 \pm 0.02$ & $0.52 \pm 0.02$ & {\color{green} $\mathbf{0.54 \pm 0.02}$} & {\color{green} $\mathbf{0.57 \pm 0.03}$} & {\color{green} $\mathbf{0.60 \pm 0.02}$} & {\color{green} $\mathbf{0.52 \pm 0.01}$} & {\color{green} $\mathbf{0.52 \pm 0.01}$} & $0.53 \pm 0.03$ & $0.54 \pm 0.03$ & $0.58 \pm 0.04$ \\
    1614153 & $0.33 \pm 0.01$ & $0.34 \pm 0.01$ & $0.33 \pm 0.01$ & $0.34 \pm 0.01$ & $0.34 \pm 0.01$ & {\color{green} $\mathbf{0.33 \pm 0.01}$} & {\color{green} $\mathbf{0.34 \pm 0.01}$} & {\color{green} $\mathbf{0.34 \pm 0.01}$} & {\color{green} $\mathbf{0.35 \pm 0.01}$} & {\color{green} $\mathbf{0.36 \pm 0.01}$} \\
    1614292 & {\color{green} $\mathbf{0.35 \pm 0.02}$} & {\color{green} $\mathbf{0.36 \pm 0.02}$} & {\color{green} $\mathbf{0.37 \pm 0.02}$} & {\color{green} $\mathbf{0.37 \pm 0.02}$} & {\color{green} $\mathbf{0.39 \pm 0.01}$} & $0.35 \pm 0.01$ & $0.35 \pm 0.01$ & $0.35 \pm 0.01$ & $0.35 \pm 0.01$ & $0.35 \pm 0.01$ \\
    1614423 & $0.44 \pm 0.07$ & {\color{green} $\mathbf{0.52 \pm 0.10}$} & {\color{green} $\mathbf{0.64 \pm 0.03}$} & {\color{green} $\mathbf{0.69 \pm 0.05}$} & {\color{green} $\mathbf{0.71 \pm 0.04}$} & {\color{green} $\mathbf{0.45 \pm 0.05}$} & $0.47 \pm 0.07$ & $0.51 \pm 0.07$ & $0.60 \pm 0.07$ & $0.68 \pm 0.02$ \\
    1614433 & {\color{green} $\mathbf{0.46 \pm 0.02}$} & $0.46 \pm 0.02$ & {\color{green} $\mathbf{0.47 \pm 0.02}$} & {\color{green} $\mathbf{0.49 \pm 0.03}$} & {\color{green} $\mathbf{0.52 \pm 0.04}$} & $0.44 \pm 0.01$ & {\color{green} $\mathbf{0.46 \pm 0.03}$} & $0.47 \pm 0.03$ & $0.48 \pm 0.04$ & $0.49 \pm 0.04$ \\
    1614466 & $0.46 \pm 0.01$ & $0.46 \pm 0.02$ & $0.47 \pm 0.02$ & $0.47 \pm 0.02$ & $0.51 \pm 0.03$ & {\color{green} $\mathbf{0.47 \pm 0.01}$} & {\color{green} $\mathbf{0.48 \pm 0.02}$} & {\color{green} $\mathbf{0.49 \pm 0.02}$} & {\color{green} $\mathbf{0.49 \pm 0.02}$} & {\color{green} $\mathbf{0.51 \pm 0.03}$} \\
    1614503 & $0.35 \pm 0.04$ & {\color{green} $\mathbf{0.36 \pm 0.06}$} & $0.37 \pm 0.04$ & $0.56 \pm 0.24$ & nan & {\color{green} $\mathbf{0.37 \pm 0.02}$} & $0.36 \pm 0.03$ & {\color{green} $\mathbf{0.38 \pm 0.04}$} & {\color{green} $\mathbf{0.57 \pm 0.20}$} & nan  \\
    1614508 & {\color{green} $\mathbf{0.70 \pm 0.12}$} & $0.77 \pm 0.13$ & {\color{green} $\mathbf{0.80 \pm 0.05}$} & $0.82 \pm 0.09$ & nan & $0.66 \pm 0.08$ & {\color{green} $\mathbf{0.79 \pm 0.07}$} & $0.77 \pm 0.09$ & {\color{green} $\mathbf{0.83 \pm 0.09}$} & nan  \\
    1614522 & {\color{green} $\mathbf{0.52 \pm 0.05}$} & {\color{green} $\mathbf{0.53 \pm 0.03}$} & {\color{green} $\mathbf{0.54 \pm 0.06}$} & {\color{green} $\mathbf{0.61 \pm 0.02}$} & {\color{green} $\mathbf{0.62 \pm 0.02}$} & $0.50 \pm 0.03$ & $0.52 \pm 0.02$ & $0.52 \pm 0.04$ & $0.56 \pm 0.03$ & $0.58 \pm 0.03$ \\
    1737951 & $0.49 \pm 0.09$ & $0.46 \pm 0.07$ & {\color{green} $\mathbf{0.61 \pm 0.03}$} & {\color{green} $\mathbf{0.67 \pm 0.06}$} & nan & {\color{green} $\mathbf{0.57 \pm 0.04}$} & {\color{green} $\mathbf{0.57 \pm 0.03}$} & $0.58 \pm 0.04$ & $0.58 \pm 0.07$ & nan  \\
    1738079 & $0.48 \pm 0.01$ & $0.47 \pm 0.01$ & $0.47 \pm 0.02$ & $0.49 \pm 0.04$ & nan & {\color{green} $\mathbf{0.52 \pm 0.01}$} & {\color{green} $\mathbf{0.51 \pm 0.01}$} & {\color{green} $\mathbf{0.51 \pm 0.03}$} & {\color{green} $\mathbf{0.52 \pm 0.03}$} & nan  \\
    1738362 & $0.31 \pm 0.03$ & $0.34 \pm 0.05$ & $0.34 \pm 0.07$ & {\color{green} $\mathbf{0.64 \pm 0.23}$} & nan & {\color{green} $\mathbf{0.37 \pm 0.01}$} & {\color{green} $\mathbf{0.38 \pm 0.03}$} & {\color{green} $\mathbf{0.40 \pm 0.04}$} & $0.46 \pm 0.19$ & nan  \\
    1738395 & $0.49 \pm 0.03$ & {\color{green} $\mathbf{0.50 \pm 0.03}$} & {\color{green} $\mathbf{0.49 \pm 0.03}$} & $0.50 \pm 0.05$ & nan & {\color{green} $\mathbf{0.50 \pm 0.03}$} & $0.48 \pm 0.03$ & $0.48 \pm 0.03$ & {\color{green} $\mathbf{0.50 \pm 0.05}$} & nan  \\
    1738485 & {\color{green} $\mathbf{0.49 \pm 0.01}$} & {\color{green} $\mathbf{0.51 \pm 0.02}$} & {\color{green} $\mathbf{0.52 \pm 0.04}$} & {\color{green} $\mathbf{0.54 \pm 0.04}$} & {\color{green} $\mathbf{0.55 \pm 0.06}$} & $0.49 \pm 0.01$ & $0.49 \pm 0.01$ & $0.48 \pm 0.01$ & $0.50 \pm 0.06$ & $0.54 \pm 0.06$ \\
    1738502 & $0.39 \pm 0.01$ & $0.40 \pm 0.02$ & $0.40 \pm 0.02$ & {\color{green} $\mathbf{0.43 \pm 0.04}$} & {\color{green} $\mathbf{0.43 \pm 0.04}$} & {\color{green} $\mathbf{0.41 \pm 0.01}$} & {\color{green} $\mathbf{0.42 \pm 0.02}$} & {\color{green} $\mathbf{0.41 \pm 0.02}$} & $0.41 \pm 0.02$ & $0.41 \pm 0.02$ \\
    1738573 & $0.51 \pm 0.02$ & $0.50 \pm 0.01$ & $0.51 \pm 0.02$ & {\color{green} $\mathbf{0.52 \pm 0.02}$} & {\color{green} $\mathbf{0.53 \pm 0.02}$} & {\color{green} $\mathbf{0.52 \pm 0.01}$} & {\color{green} $\mathbf{0.51 \pm 0.01}$} & {\color{green} $\mathbf{0.52 \pm 0.01}$} & $0.52 \pm 0.01$ & $0.52 \pm 0.01$ \\
    1738579 & {\color{green} $\mathbf{0.54 \pm 0.03}$} & {\color{green} $\mathbf{0.56 \pm 0.02}$} & {\color{green} $\mathbf{0.56 \pm 0.03}$} & {\color{green} $\mathbf{0.60 \pm 0.05}$} & nan & $0.53 \pm 0.03$ & $0.55 \pm 0.03$ & $0.55 \pm 0.04$ & $0.58 \pm 0.04$ & nan  \\
    1738633 & $0.55 \pm 0.04$ & $0.56 \pm 0.04$ & $0.60 \pm 0.04$ & $0.57 \pm 0.09$ & nan & {\color{green} $\mathbf{0.59 \pm 0.02}$} & {\color{green} $\mathbf{0.60 \pm 0.02}$} & {\color{green} $\mathbf{0.60 \pm 0.04}$} & {\color{green} $\mathbf{0.63 \pm 0.12}$} & nan  \\
    1794324 & {\color{green} $\mathbf{0.52 \pm 0.02}$} & {\color{green} $\mathbf{0.52 \pm 0.01}$} & {\color{green} $\mathbf{0.52 \pm 0.01}$} & {\color{green} $\mathbf{0.53 \pm 0.02}$} & {\color{green} $\mathbf{0.54 \pm 0.02}$} & $0.51 \pm 0.00$ & $0.51 \pm 0.01$ & $0.52 \pm 0.01$ & $0.52 \pm 0.01$ & $0.53 \pm 0.01$ \\
    1794504 & {\color{green} $\mathbf{0.68 \pm 0.01}$} & {\color{green} $\mathbf{0.70 \pm 0.04}$} & {\color{green} $\mathbf{0.71 \pm 0.04}$} & {\color{green} $\mathbf{0.85 \pm 0.24}$} & nan & $0.65 \pm 0.05$ & $0.66 \pm 0.05$ & $0.71 \pm 0.04$ & {\color{green} $\mathbf{0.85 \pm 0.24}$} & nan  \\
    1794519 & $0.64 \pm 0.05$ & {\color{green} $\mathbf{0.66 \pm 0.07}$} & {\color{green} $\mathbf{0.69 \pm 0.08}$} & {\color{green} $\mathbf{0.73 \pm 0.13}$} & nan & {\color{green} $\mathbf{0.64 \pm 0.07}$} & $0.62 \pm 0.05$ & $0.65 \pm 0.07$ & $0.71 \pm 0.11$ & nan  \\
    1794557 & $0.53 \pm 0.02$ & $0.54 \pm 0.01$ & $0.54 \pm 0.02$ & $0.53 \pm 0.02$ & $0.53 \pm 0.01$ & {\color{green} $\mathbf{0.55 \pm 0.01}$} & {\color{green} $\mathbf{0.55 \pm 0.01}$} & {\color{green} $\mathbf{0.55 \pm 0.01}$} & {\color{green} $\mathbf{0.55 \pm 0.01}$} & {\color{green} $\mathbf{0.55 \pm 0.01}$} \\
    1963701 & {\color{green} $\mathbf{0.60 \pm 0.01}$} & {\color{green} $\mathbf{0.59 \pm 0.02}$} & {\color{green} $\mathbf{0.59 \pm 0.02}$} & {\color{green} $\mathbf{0.60 \pm 0.01}$} & {\color{green} $\mathbf{0.60 \pm 0.02}$} & $0.60 \pm 0.01$ & $0.59 \pm 0.01$ & {\color{green} $\mathbf{0.59 \pm 0.01}$} & $0.59 \pm 0.01$ & $0.59 \pm 0.02$ \\
    1963705 & $0.71 \pm 0.03$ & $0.70 \pm 0.02$ & $0.72 \pm 0.01$ & $0.72 \pm 0.01$ & $0.71 \pm 0.02$ & {\color{green} $\mathbf{0.77 \pm 0.01}$} & {\color{green} $\mathbf{0.77 \pm 0.02}$} & {\color{green} $\mathbf{0.77 \pm 0.01}$} & {\color{green} $\mathbf{0.77 \pm 0.03}$} & {\color{green} $\mathbf{0.76 \pm 0.04}$} \\
    1963715 & {\color{green} $\mathbf{0.63 \pm 0.00}$} & {\color{green} $\mathbf{0.62 \pm 0.03}$} & $0.63 \pm 0.01$ & $0.64 \pm 0.01$ & $0.64 \pm 0.02$ & $0.63 \pm 0.01$ & $0.61 \pm 0.04$ & {\color{green} $\mathbf{0.64 \pm 0.01}$} & {\color{green} $\mathbf{0.64 \pm 0.01}$} & {\color{green} $\mathbf{0.64 \pm 0.02}$} \\
    1963721 & $0.50 \pm 0.02$ & $0.50 \pm 0.02$ & $0.50 \pm 0.02$ & {\color{green} $\mathbf{0.51 \pm 0.04}$} & {\color{green} $\mathbf{0.55 \pm 0.07}$} & {\color{green} $\mathbf{0.52 \pm 0.02}$} & {\color{green} $\mathbf{0.51 \pm 0.01}$} & {\color{green} $\mathbf{0.52 \pm 0.01}$} & $0.51 \pm 0.03$ & $0.54 \pm 0.09$ \\
    1963723 & $0.72 \pm 0.03$ & $0.73 \pm 0.01$ & $0.73 \pm 0.01$ & $0.74 \pm 0.02$ & $0.75 \pm 0.02$ & {\color{green} $\mathbf{0.72 \pm 0.03}$} & {\color{green} $\mathbf{0.74 \pm 0.02}$} & {\color{green} $\mathbf{0.74 \pm 0.01}$} & {\color{green} $\mathbf{0.75 \pm 0.02}$} & {\color{green} $\mathbf{0.77 \pm 0.02}$} \\
    1963731 & $0.70 \pm 0.02$ & $0.70 \pm 0.01$ & $0.70 \pm 0.02$ & $0.76 \pm 0.06$ & $0.76 \pm 0.05$ & {\color{green} $\mathbf{0.71 \pm 0.03}$} & {\color{green} $\mathbf{0.74 \pm 0.02}$} & {\color{green} $\mathbf{0.74 \pm 0.02}$} & {\color{green} $\mathbf{0.76 \pm 0.02}$} & {\color{green} $\mathbf{0.79 \pm 0.03}$} \\
    1963741 & {\color{green} $\mathbf{0.65 \pm 0.02}$} & {\color{green} $\mathbf{0.66 \pm 0.00}$} & {\color{green} $\mathbf{0.66 \pm 0.01}$} & {\color{green} $\mathbf{0.67 \pm 0.01}$} & {\color{green} $\mathbf{0.67 \pm 0.01}$} & $0.65 \pm 0.01$ & $0.65 \pm 0.01$ & $0.66 \pm 0.01$ & $0.65 \pm 0.01$ & $0.65 \pm 0.01$ \\
    1963756 & {\color{green} $\mathbf{0.76 \pm 0.05}$} & {\color{green} $\mathbf{0.75 \pm 0.04}$} & {\color{green} $\mathbf{0.76 \pm 0.04}$} & {\color{green} $\mathbf{0.77 \pm 0.02}$} & {\color{green} $\mathbf{0.77 \pm 0.01}$} & $0.71 \pm 0.02$ & $0.72 \pm 0.03$ & $0.70 \pm 0.04$ & $0.72 \pm 0.02$ & $0.72 \pm 0.03$ \\
    1963773 & $0.60 \pm 0.01$ & $0.59 \pm 0.04$ & $0.60 \pm 0.03$ & $0.60 \pm 0.02$ & $0.60 \pm 0.02$ & {\color{green} $\mathbf{0.62 \pm 0.02}$} & {\color{green} $\mathbf{0.63 \pm 0.02}$} & {\color{green} $\mathbf{0.63 \pm 0.02}$} & {\color{green} $\mathbf{0.64 \pm 0.03}$} & {\color{green} $\mathbf{0.63 \pm 0.02}$} \\
    1963799 & $0.63 \pm 0.01$ & $0.62 \pm 0.01$ & $0.61 \pm 0.03$ & $0.61 \pm 0.02$ & $0.60 \pm 0.05$ & {\color{green} $\mathbf{0.64 \pm 0.02}$} & {\color{green} $\mathbf{0.66 \pm 0.01}$} & {\color{green} $\mathbf{0.66 \pm 0.02}$} & {\color{green} $\mathbf{0.67 \pm 0.03}$} & {\color{green} $\mathbf{0.67 \pm 0.07}$} \\
    1963810 & {\color{green} $\mathbf{0.78 \pm 0.01}$} & {\color{green} $\mathbf{0.77 \pm 0.02}$} & {\color{green} $\mathbf{0.78 \pm 0.01}$} & {\color{green} $\mathbf{0.78 \pm 0.01}$} & {\color{green} $\mathbf{0.78 \pm 0.01}$} & $0.77 \pm 0.01$ & $0.75 \pm 0.03$ & $0.76 \pm 0.02$ & $0.76 \pm 0.02$ & $0.78 \pm 0.02$ \\
    1963818 & {\color{green} $\mathbf{0.71 \pm 0.02}$} & {\color{green} $\mathbf{0.72 \pm 0.01}$} & $0.72 \pm 0.03$ & {\color{green} $\mathbf{0.73 \pm 0.02}$} & {\color{green} $\mathbf{0.73 \pm 0.03}$} & $0.70 \pm 0.04$ & $0.72 \pm 0.01$ & {\color{green} $\mathbf{0.72 \pm 0.03}$} & $0.72 \pm 0.03$ & $0.73 \pm 0.03$ \\
    1963819 & $0.59 \pm 0.04$ & $0.61 \pm 0.03$ & $0.59 \pm 0.01$ & $0.61 \pm 0.03$ & $0.61 \pm 0.03$ & {\color{green} $\mathbf{0.60 \pm 0.02}$} & {\color{green} $\mathbf{0.61 \pm 0.03}$} & {\color{green} $\mathbf{0.62 \pm 0.03}$} & {\color{green} $\mathbf{0.63 \pm 0.04}$} & {\color{green} $\mathbf{0.63 \pm 0.03}$} \\
    1963824 & {\color{green} $\mathbf{0.62 \pm 0.04}$} & $0.63 \pm 0.02$ & {\color{green} $\mathbf{0.65 \pm 0.04}$} & {\color{green} $\mathbf{0.65 \pm 0.04}$} & {\color{green} $\mathbf{0.69 \pm 0.05}$} & $0.62 \pm 0.03$ & {\color{green} $\mathbf{0.64 \pm 0.04}$} & $0.64 \pm 0.04$ & $0.63 \pm 0.03$ & $0.63 \pm 0.04$ \\
    1963825 & $0.73 \pm 0.01$ & $0.72 \pm 0.02$ & $0.73 \pm 0.01$ & $0.74 \pm 0.02$ & $0.75 \pm 0.04$ & {\color{green} $\mathbf{0.77 \pm 0.02}$} & {\color{green} $\mathbf{0.77 \pm 0.02}$} & {\color{green} $\mathbf{0.79 \pm 0.02}$} & {\color{green} $\mathbf{0.79 \pm 0.03}$} & {\color{green} $\mathbf{0.80 \pm 0.04}$} \\
    1963827 & $0.78 \pm 0.03$ & {\color{green} $\mathbf{0.80 \pm 0.01}$} & {\color{green} $\mathbf{0.80 \pm 0.03}$} & {\color{green} $\mathbf{0.79 \pm 0.02}$} & {\color{green} $\mathbf{0.80 \pm 0.02}$} & {\color{green} $\mathbf{0.78 \pm 0.01}$} & $0.76 \pm 0.04$ & $0.76 \pm 0.03$ & $0.77 \pm 0.01$ & $0.76 \pm 0.02$ \\
    1963831 & {\color{green} $\mathbf{0.69 \pm 0.01}$} & {\color{green} $\mathbf{0.69 \pm 0.02}$} & $0.69 \pm 0.01$ & $0.69 \pm 0.02$ & $0.68 \pm 0.06$ & $0.68 \pm 0.02$ & $0.68 \pm 0.02$ & {\color{green} $\mathbf{0.70 \pm 0.02}$} & {\color{green} $\mathbf{0.71 \pm 0.03}$} & {\color{green} $\mathbf{0.72 \pm 0.05}$} \\
    1963838 & $0.49 \pm 0.01$ & $0.49 \pm 0.01$ & $0.51 \pm 0.04$ & {\color{green} $\mathbf{0.55 \pm 0.03}$} & nan & {\color{green} $\mathbf{0.54 \pm 0.02}$} & {\color{green} $\mathbf{0.53 \pm 0.01}$} & {\color{green} $\mathbf{0.55 \pm 0.02}$} & $0.52 \pm 0.04$ & nan  \\
        \bottomrule
    \end{tabular}
}
\vspace{-0.2cm}
\label{table:50_maml_fsmol}
\end{table*}

\end{document}